%% file: main.tex
\documentclass{article}

\usepackage[final]{neurips_2025}

\usepackage[utf8]{inputenc} 
\usepackage[T1]{fontenc}    
\usepackage{hyperref}       
\usepackage{url}            
\usepackage{booktabs}       
\usepackage{amsfonts}       
\usepackage{nicefrac}       
\usepackage{microtype}      
\usepackage{xcolor}         

\input{preamble}

\title{
\vspace{-0.6em}
\ours~\includegraphics[height=0.8cm]{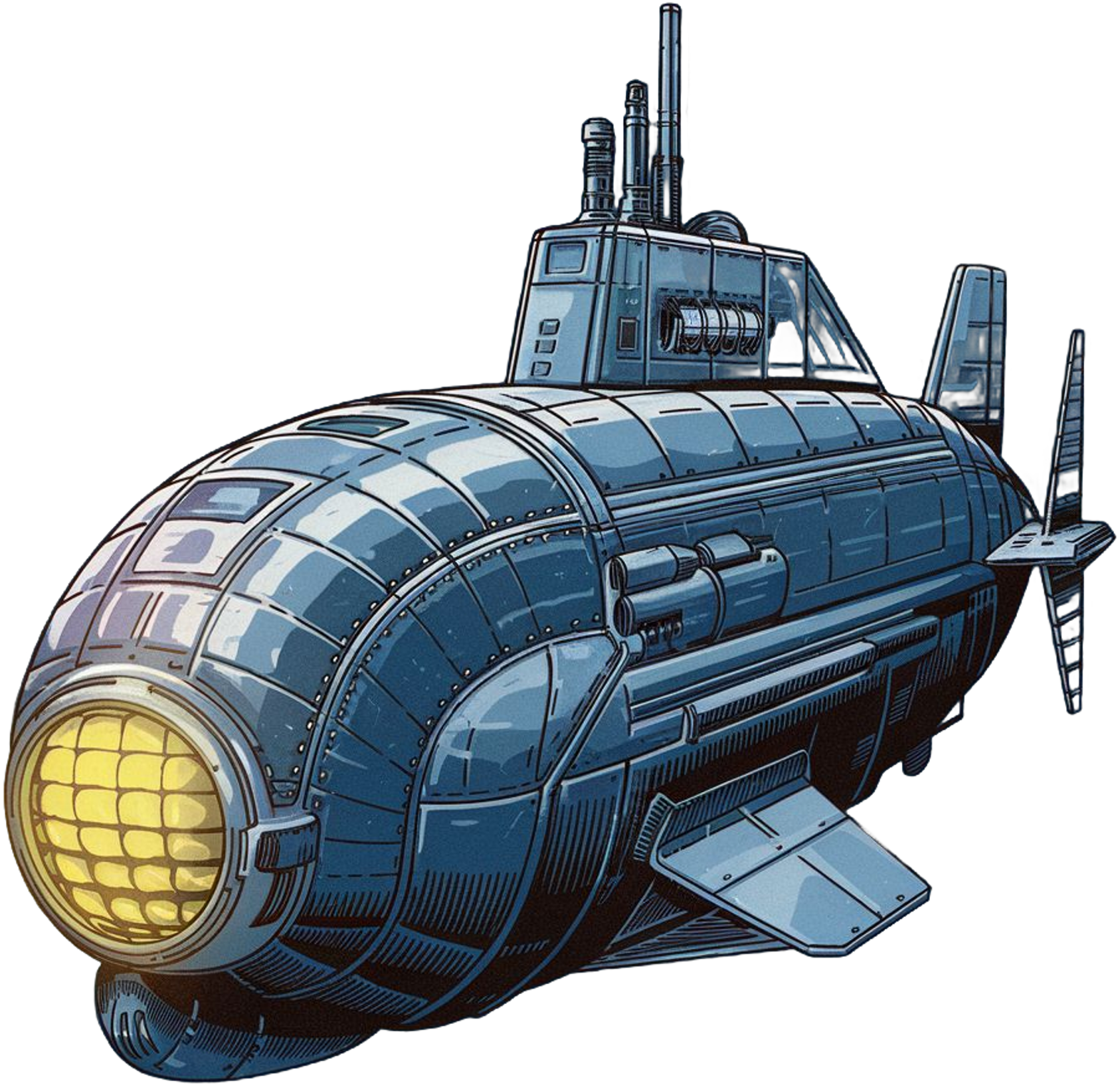}: A Large Multimodal Model for Underwater Scene Understanding
\vspace{-0.3em}
}

\renewcommand\footnotemark{}
\author{Wei Xu$^{1\ast}$, Cheng Wang$^{1\ast}$, Dingkang Liang$^{1}$, Zongchuang Zhao$^{1}$, \\ \textbf{Xingyu Jiang$^{1}$, Peng Zhang$^{2}$, Xiang Bai$^{1\dagger}$}\thanks{\footnotesize{$^\ast$Equal contribution. $^\dagger$Corresponding author.}}\\
	$^1$Huazhong University of Science and Technology
	\\
    $^2$National University of Defense Technology
    \\
    \texttt{\{wxu2023, cwang666, dkliang, xbai\}@hust.edu.cn}
}

\begin{document}

\maketitle

{%
\begin{figure}[H]
\hsize=\textwidth
\centering
\includegraphics[width=1.0\linewidth]{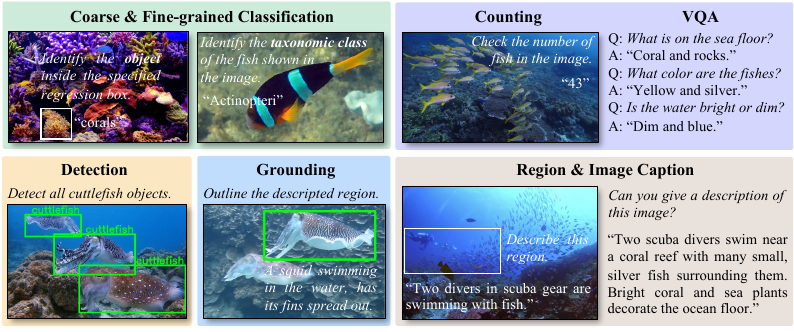}
\caption{
The underwater environment presents a visually rich and dynamically evolving landscape. 
\ours~addresses eight diverse underwater tasks, encompassing coarse-grained classification, fine-grained classification, counting, visual question answering (VQA), detection, grounding, region caption, and image caption, enabling comprehensive understandings across multiple granularities.
}
\label{fig:intro}
\end{figure}
}

\input{sections/abstract}

\input{sections/introduction}

\input{sections/relatedwork}

\input{sections/dataset}

\input{sections/methodology}
\input{sections/experiments}

\input{sections/conclusion}

{\small
\bibliographystyle{plain}
\bibliography{main}
}

\input{sections/checklist}

\input{sections/appendix}

\end{document}

%% file: preamble.tex
\def\ours{\textsc{Nautilus}}
\def\oursdata{NautData}
\def\oursbench{NautData test set}
\def\datasum{$1.45$}

\usepackage{multirow}
\usepackage{float}
\usepackage{pifont}
\usepackage{bm}
\definecolor{darkgreen}{rgb}{0, 0.6, 0}
\definecolor{darkred}{rgb}{0.8, 0, 0}
\newcommand{\cmark}{\ding{51}}%
\newcommand{\greencmark}{\textcolor{darkgreen}{\cmark}}

\newcommand{\modelvariant}[1]{\textcolor{gray}{\scalebox{0.75}{#1}}}

\usepackage{algorithm}
\usepackage{algorithmic}
\usepackage{listings}
\usepackage{graphicx}
\usepackage{enumitem}
\usepackage{xspace}
\usepackage{wrapfig}
\usepackage{caption}
\usepackage{colortbl}
\usepackage{hyperref}
\usepackage{tcolorbox}
\usepackage{amsmath}
\usepackage{threeparttable}

\newcommand{\mysection}[1]{\noindent\textbf{#1.}}
\newcommand{\PromptSty}[1]{\textnormal{\color{blue!90!black}#1}\unskip}

\definecolor{isabelline}{RGB}{244, 240, 236}
\definecolor{kaiming-green}{RGB}{57,181,74} 
\hypersetup{
    colorlinks=true,
    linkcolor=red,
    filecolor=magenta,      
    urlcolor=magenta,
    citecolor=kaiming-green
}

%% file: sections/abstract.tex
\begin{abstract}

Underwater exploration offers critical insights into our planet and attracts increasing attention for its broader applications in resource exploration, national security, etc. We study the underwater scene understanding methods, which aim to achieve automated underwater exploration. The underwater scene understanding task demands multi-task perceptions from multiple granularities. However, the absence of large-scale underwater multi-task instruction-tuning datasets hinders the progress of this research. To bridge this gap, we construct \oursdata, a dataset containing \datasum\,M image-text pairs supporting eight underwater scene understanding tasks. It enables the development and thorough evaluation of the underwater scene understanding models. Underwater image degradation is a widely recognized challenge that interferes with underwater tasks. To improve the robustness of underwater scene understanding, we introduce physical priors derived from underwater imaging models and propose a plug-and-play vision feature enhancement (VFE) module, which explicitly restores clear underwater information. We integrate this module into renowned baselines LLaVA-1.5 and Qwen2.5-VL and build our underwater LMM, \ours. Experiments conducted on the \oursdata~and public underwater datasets demonstrate the effectiveness of the VFE module, consistently improving the performance of both baselines on the majority of supported tasks, thus ensuring the superiority of \ours~in the underwater scene understanding area. Data and models are available at \url{https://github.com/H-EmbodVis/NAUTILUS}.

\end{abstract}

%% file: sections/introduction.tex
\section{Introduction}
\label{sec:intro}

The underwater world plays a pivotal role in global well-being, as it covers more than $70$\% of the Earth's surface and encompasses the largest ecosystem on our planet~\cite{danovaro2025assessing, beauchesne2025ecological,nowakowski2023co,ramirez2010deep}.
Advancing underwater scene understanding methods facilitates automated underwater robot exploration~\cite{skaugset2025autonomous}, benefiting sufficient environmental protection~\cite{jacquemont20243d} and resource development~\cite{yoerger2021hybrid}.
Comprehensive underwater scene understanding comprises both perception (e.g., detection and counting) and semantic understanding tasks (e.g., region and image caption).
However, most underwater methods~\cite{zhou2024amsp,sun2023indiscernible,zhang2021self} are typically tailored to a specific task, limiting their understanding of underwater scenes to a task-specific perspective. 

Recent achievements in the general domain have driven the advancement of specialized LMMs in understanding tasks~\cite{zheng2024video,wang2025embodied}, leading to prominent applications in fields like autonomous driving~\cite{pan2024vlp}, embodied intelligence~\cite{driess2023palm}, and document understanding~\cite{li2024monkey}.
The promising abilities of large multimodal models (LMMs) provide a considerable solution for the underwater scene understanding area.
Nonetheless, we empirically find that \textit{directly adopting general LMMs in underwater scenes cannot serve as an ideal solution}, as they face inherent challenges arising from 1) the significant domain shift between in-air and underwater data, and 2) image degradation due to light scattering and absorption in water.

OceanGPT~\cite{bi2023oceangpt} infuses the richness of knowledge into a large language model (LLM), uncovering its potential in the field of ocean science while failing to interpret multimodal inputs. 
MarineGPT~\cite{zheng2023marinegpt} empowers the LLMs to sense vision-language information and has been the first-ever publicly available underwater LMM. As shown in Tab.~\ref{tab:datasets}, this pioneering work constructs a large-scale underwater vision-language dataset for instruction tuning while only focusing on the image-level understanding, neglecting the hierarchical underwater scene information. We take a valuable step in constructing \oursdata, an underwater instruction-following dataset containing \datasum~million question-answer pairs that cover eight diverse underwater tasks.
It establishes a solid foundation for the development of underwater LMMs, bypassing the in-air domain shift prevalent in most current instruction-tuning datasets.

The image degradation problem hinders reliable underwater scene understanding. Most methods leave this challenge to the training process, driving the models to learn underwater representations on their own, which could be sub-optimal due to the intricate complexity of underwater conditions. 
To bridge this gap, we propose a plug-and-play vision feature enhancement (VFE) module that explicitly removes noise responses introduced by image degradation and enhances the understanding performance of underwater LMMs. 
Specifically, the underwater imaging model provides a physical representation of image degradation in underwater scenarios~\cite{zhou2023underwater,zhang2022underwater,akkaynak2018revised,nathan2024osmosis}, emphasizing backscattering from the surroundings as a primary interference. We adopt a dark pixel prior to quantifying the intensity of backscattering, paving the way for removing its adverse effects. The optical medium influences imaging quality, with underwater imaging facing significant light absorption compared to in-air imaging, and is another crucial factor contributing to underwater image degradation. To address this, we extract depth information to restore the scene signals attenuated by medium absorption. The VFE module can be flexibly employed in general LMMs, and we integrate it into two renowned baselines, LLaVA-1.5~\cite{liu2024improved} and Qwen2.5-VL~\cite{bai2025qwen2}, to build our underwater LMM \ours.

We evaluate the \ours~and prestigious LMMs~\cite{liu2024improved,ye2024mplug3,chen2024expanding,bai2025qwen2} on the \oursdata~to analyze their underwater scene understanding performance in a supervised manner. Then, we directly evaluate them on MarineInst~\cite{zheng2024marineinst}, a recent large-scale vision-language underwater dataset. This zero-shot experiment indicates the generalization capabilities of our method. 

This work contributes to three aspects: \textbf{1)} We construct \oursdata, a large-scale underwater instruction-following dataset containing \datasum\,M image-text pairs, enabling developments and evaluations of underwater LMMs. \textbf{2)} We build the first eight-task underwater LMM \ours, achieving underwater scene understanding from image, region, and object levels. It empowers comprehensive underwater scene understanding through aggregating hierarchical scene information. \textbf{3)} We design a plug-and-play VFE module motivated by a physical underwater imaging model. It restores degraded information explicitly in the feature space. Experiments on renowned baselines demonstrate its effectiveness on all the annotated tasks.

%% file: sections/relatedwork.tex
\section{Related Work}

\input{tables/datasets}

\textbf{Underwater Vision-Language Analysis.}
Pioneering studies~\cite{liu2023visual,zhu2023minigpt,liu2024improved} have aligned linguistic and visual representations using projection layers, enabling LLMs to process multimodal information and advance into LMMs. Through visual instruction tuning, LMMs have presented impressive performance in
both general-purpose and domain-specific areas. 
For instance, LLaVA-NeXT~\cite{liu2024llavanext}, InternVL~\cite{chen2024internvl}, and MiniGPT-v2~\cite{chen2023minigpt} have achieved remarkable success in the general domain by advancing LMMs through improvements in data, vision encoder structures, and training strategies.
PaLM-E~\cite{driess2023palm} and Dolphins~\cite{ma2024dolphins} find LMMs effective in embodied intelligence and autonomous driving, respectively.
MarineGPT~\cite{zheng2023marinegpt} trains LMMs with marine science knowledge, focusing primarily on image-level understanding while lacking attention to region- and object-level~\cite{muhtar2024lhrs,liang2025sood++} scene information.
AquaticCLIP~\cite{alawode2025aquaticclip} employs contrastive learning-based pretraining on aquatic image-text pairs to align the underwater image and text representations. The AquaticCLIP enhances performance on downstream underwater tasks while lacking the capability to follow instructions directly.
MarineInst~\cite{zheng2024marineinst} localizes objects and generates linguistic descriptions for each of them.
The object-level vision and text responses contribute to detailed marine image analysis. CoralSCOP~\cite{zheng2024coralscop} can be driven by both vision and text prompts, enabling the mask generation of corals described by users. 
Despite increasing attention, research on underwater LMMs remains limited and requires further efforts to achieve underwater vision-language dialogues. This paper advances this field by exploring the capabilities of LLMs to deliver hierarchical underwater scene understandings.

\textbf{Underwater Image Enhancement.}
Underwater images often exhibit poor visibility, low contrast, and severe color distortions, primarily due to light absorption and scattering in aquatic environments. To address these issues, underwater image enhancement methods aim to mitigate such degradations and restore visual quality comparable to in-air images. Conventional methods employ typical image augmentation strategies, such as histogram stretching~\cite{dong2022underwater,zhang2021color} and image fusion~\cite{guo2020multi} to enhance the underwater images.
These methods are easy to deploy, but they suffer from the generalization limitations of handcrafted features.
Another direction of underwater image enhancement research~\cite{fu2024multi,wu2022fw, fabbri2018enhancing} explores deep learning-based methods, particularly those leveraging generative adversarial networks (GANs) to improve the quality of underwater images. However, these methods are hindered by the inadequate availability of high-quality training data, as collecting underwater images is often constrained by the complexity of underwater environments, high equipment costs, and the challenges associated with accurate data annotation. Physical model-based methods~\cite{zhou2023underwater,zhang2022underwater,tong2022towards} offer a feasible solution to these problems. They reduce the search space for parameters by incorporating handcrafted priors, thus mitigating dependency on large-scale training data. 
However, directly applying image enhancement to underwater images may result in information loss, thereby limiting the effectiveness in underwater scene understanding, as demonstrated in our experiments.
Building upon the previous discussion, we innovatively introduce an enhancement in feature space within an LMM, providing an efficient solution to extract underwater visual information.

%% file: tables/datasets.tex
\setlength\tabcolsep{1.2pt}
\begin{table*}
\scriptsize
\setlength{\baselineskip}{1.05\baselineskip}
\begin{threeparttable}
\caption{The comparisons between \oursdata~and recent underwater vision-language datasets. Our dataset is more comprehensive for eight-task annotations involving understandings at three granularities, making it a valuable contribution to the community.}
\begin{tabular*}{\linewidth}{@{\extracolsep{\fill}} lcccccccccccccc }
\toprule
\multirow{4}{*}{Datasets} &\multirow{4}{*}{Reference} & \multicolumn{8}{c}{Supported tasks} & \multicolumn{3}{c}{Granularity} & \multirow{4}{*}{\shortstack{QA\\pairs}}  & \multirow{4}{*}{\shortstack{Open\\source}}  \\ 
\cmidrule{3-10}\cmidrule{11-13}
 & & \multirow{2}{*}{VQA} & \multirow{2}{*}{Detection} & \multicolumn{2}{c}{Classification}  & \multirow{2}{*}{Grounding} & \multicolumn{2}{c}{Caption} & \multirow{2}{*}{Counting} & \multirow{2}{*}{Img.} & \multirow{2}{*}{Reg.} & \multirow{2}{*}{Obj.} &  \\
\cmidrule{5-6}\cmidrule{8-9}
&&&&Coarse&Fine&&Image&Region&&&&&& \\

\midrule

MarineGPT~\cite{zheng2023marinegpt} & arXiv 23 & - & - & - & - & - & \greencmark & - & - & \greencmark & - & - & $1.12$ M & Not Avail. \\
MarineInst20M~\cite{zheng2024marineinst} & ECCV 24 & -  & \greencmark & - & - & \greencmark & \greencmark & - & - & \greencmark & - & \greencmark & $20$ M & Part. Avail.\tnote{a}\\ 
CoralMask~\cite{zheng2024coralscop} & CVPR 24 & -  & \greencmark & - & - & \greencmark & - & - & - & - & - & \greencmark & $46.61$ K & Avail. \\
AquaticCLIP~\cite{alawode2025aquaticclip} & arXiv 25 & -  & - & - & - & - & \greencmark & - & - & \greencmark & - & -  & $2$ M & Not Avail. \\
SynTIDE~\cite{lin2025tide} & CVPR 25 & -  & - & - & - & - & \greencmark & - & - & \greencmark & - & - & $54.51$ K & Avail. \\
\midrule
\oursdata~(\textbf{ours}) & - & \greencmark & \greencmark & \greencmark & \greencmark & \greencmark & \greencmark & \greencmark & \greencmark & \greencmark & \greencmark & \greencmark & \datasum\,M & Avail. \\
\bottomrule
\end{tabular*}
\begin{tablenotes}
  \footnotesize
  \item[a] Current public version contains $2.2$\,M QA pairs.
\end{tablenotes}
\label{tab:datasets}
\end{threeparttable}
\end{table*}

%% file: sections/dataset.tex
\section{Dataset Construction}

\input{figures/construct}

Underwater vision-language datasets~\cite{zheng2023marinegpt, zheng2024marineinst} lack multi-granular and multi-task annotations. To address this gap, we construct \oursdata, which provides a pioneering resource for advancing research on underwater scene understanding.
The distinct advantages of \oursdata~lie in three aspects: (1) extensive underwater information encompassing image-, region-, and object-level understanding; (2) diverse conversational structures, including both rule-based and free-form content; and (3) significant scale, comprising $158$\,K images and \datasum\,M QA pairs.

\textbf{Attributes of Individual Entities.} Public underwater object detection datasets such as RUOD~\cite{fu2023rethinking}, Deepfish~\cite{saleh2020realistic}, and Brackish~\cite{pedersen2019detection} provide bounding-box annotations reflecting the position information of entities. As shown in Fig.~\ref{fig:construct}, we employ a rule-based procedure to reformulate the labeled coordinates into a linguistic answer ``\texttt{Class:[$x_1, y_1, x_2, y_2$],..., Class:[$x_1, y_1, x_2, y_2$]}''. Then, we concatenate it with a pre-defined question ``\texttt{Detect all underwater objects in the image.}'' to create an object-level perception conversation.

Classifying aquatic targets constitutes an expert-level task due to the specialized nature of underwater knowledge. We design both coarse-grained and fine-grained classification conversations, facilitating knowledge sharing in this domain.
Specifically, a standard detection task distinguishes underwater objects in coarse-grained categories, e.g., fish, turtles, or reefs. FishNet~\cite{khan2023fishnet} further fine-grains the fish into $8$ taxonomic classes. 
We pre-define questions with corresponding detection categories as answers to deliver coarse-grained classification conversations. Subsequently, for the FishNet dataset, we couple questions with ground-truth taxonomic class names as answers to construct fine-grained classification dialogs.

Apart from the formalized annotations of positions and categories, we employ LMMs~\cite{team2024gemini,bai2025qwen2,openai2024chatml} to generate textual descriptions, which could contain rich individual properties, such as color, texture, and shape. 
For instance, we inject object coordinates in textual form into prompts, directing concentrations of LMMs on target local regions. The generated descriptions are subsequently integrated with these coordinates to construct grounding conversations.

\textbf{Annotations on Regional Groups.}
Underwater species often exhibit collective behaviors essential for understanding their survival and ecosystem dynamics. IOCfish5K~\cite{sun2023indiscernible} is a densely distributed underwater object counting dataset, averaging $117$ targets per image. We design conversations on group counts and behaviors to enhance regional understanding. Specifically, we first couple pre-defined questions with ground-truth counts as answers, prompting models to regress numerical results directly. We then convert this regression task into a single-choice question by randomly selecting intervals from $\{5, 50, 100\}$ to construct four-term arithmetic sequences that include the ground-truth count. Additionally, we employ LMMs~\cite{team2024gemini,bai2025qwen2,openai2024chatml} to generate descriptions for regional groups, which may describe collective behaviors and relationships. These descriptions are integrated with pre-defined questions to create region caption dialogs.

\textbf{Descriptions for Holistic Semantics.}
Image-caption pairs have been demonstrated to significantly benefit LMMs in aligning vision and language modalities~\cite{bai2025qwen2,liu2023visual,kar2024brave}. 
We feed collected images into LMMs~\cite{team2024gemini,bai2025qwen2,openai2024chatml} to obtain a holistic understanding of the given scenes. The outputs are coupled with pre-defined questions to construct image caption conversations. Underwater images record rich information rarely discussed in existing datasets, such as brightness and geomorphology. We employ LMMs~\cite{team2024gemini,bai2025qwen2,openai2024chatml} to generate free-form visual question-answering (VQA) conversations discussing these elements.

During the generation process utilizing LMMs, we first employ Gemini 2.0 Flash to produce initial outputs. Subsequently, these outputs are evaluated by Qwen2.5-VL-72B, and any responses identified as low-quality are replaced with newly generated answers. Leveraging the constructed \oursdata, we collect image-text pairs to develop the \oursbench, which is further assessed by GPT-4o. Answers flagged as low-quality undergo additional manual verification by our research team.

The current \oursbench~comprises $3,920$ images paired with $7,916$ question-answering (QA) examples. These QA pairs encompass a diverse range of tasks, including image and region captioning, coarse-grained and fine-grained classification, grounding, detection, counting, and visual question answering (VQA). This comprehensive benchmark is designed to facilitate a rigorous evaluation of methods in underwater scene understanding, thereby promoting further advancements in this field.

%% file: figures/construct.tex
\begin{figure*}[t] 
\centering
\includegraphics[width=1.0\textwidth]{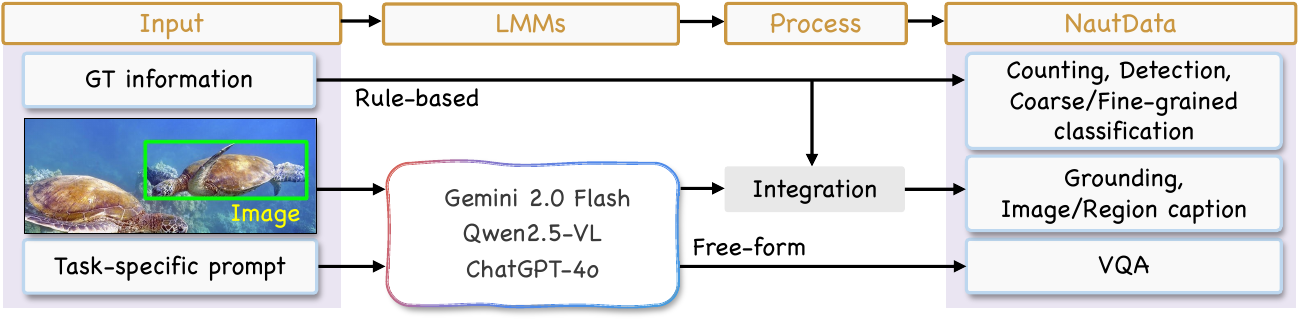}
\caption{Illustration of the data construction framework. Eight tasks are involved, and the data generation process is tailored to each task. Rule-based generation utilizes predefined templates to generate question-answer pairs. Integration generation integrates question-answer pairs using both templates and outputs from LMMs. Free-form generation enables LMMs to construct questions and answers based on the content they focus on.
}
\label{fig:construct}
\end{figure*}

%% file: sections/methodology.tex
\section{Methodology}

Underwater environments, characterized by intricate conditions and entities of diverse colors, shapes, and scales, require multi-grained perception to achieve comprehensive scene understanding.
\ours~is the first model empowering vision-language conversations spanning image-, region-, and object-level underwater scene understanding tasks, potentially facilitating seamless human-computer interaction and underwater knowledge sharing.

To illustrate the design of \ours, we first review the physical underwater imaging model to depict the motivation of explicitly dealing with underwater image degradation. Then, we detail the implementation of the model architecture involving the overall framework and the vision feature enhancement (VFE) module.

\subsection{Preliminaries}
\label{sec:Preliminaries}
Underwater imaging model~\cite{zhou2023underwater,zhang2022underwater,akkaynak2018revised,nathan2024osmosis} typically formulates the captured underwater image $\bm{I}_{c}$ as the combination of the direct reflection $\bm{D}_{c}$ from the underwater subjects and the backscattering $\bm{B}_{c}$ from the surroundings:
\begin{equation}
\bm{I}_{c} = \bm{D}_{c}+\bm{B}_{c},\quad\bm{D}_{c} = \bm{J}_{c}e^{-\beta _{c}\left ( \bm{z} \right ) \cdot \bm{z}},
\label{eq:imaging}
\end{equation}
where $\bm{J}_{c}$ is the original color at the $c$-th channel without light absorption during propagation. The underwater imaging model assumes one attenuation coefficient $e^{-\beta _{c}\left ( \bm{z} \right ) \cdot \bm{z}}$ for each color, decreasing exponentially with the imaging distance $\bm{z}$. The $\beta _{c}\left ( \bm{z} \right )$ is an unavailable parameter related to data collection conditions, also primarily regulated by imaging depth. 
Inspired by this physical imaging model, we attempt to restore the representations of $\bm{J}_{c}$ as enhanced vision features:
\begin{equation}
\bm{J}_{c}=\frac{\left ( \bm{I}_{c}-\bm{B}_{c} \right )  }{e^{-\beta _{c}\left ( \bm{z} \right ) \cdot \bm{z}}} .
\label{eq:Jc}
\end{equation}
In practice, we fit the attenuation coefficient by employing a learning module with depth information as inputs. Furthermore, we introduce a dark pixel prior~\cite{zhou2023underwater} to quantify the impact of backscattering $\bm{B}_{c}$. For instance, due to the surrounding backscattering in underwater environments, dark pixels often lose their original black appearance and instead exhibit a blue-green color. 
The dark pixel prior emphasizes that these distorted pixel values located at dark pixels reflect influences of backscatterings.
Drawing on this analysis, we localize the dark pixels of a given image and regard their responses as backscattering intensities. Subsequently, simply substitute the obtained parameters in Eq.~\ref{eq:Jc} to complete the feature enhancement.

The underwater imaging model offers a physical principle to regularize the learning phase, which we consider an explainable and sufficient optimization direction for model design.

\input{figures/pipeline}

\subsection{Model Architecture}
As shown in Fig.~\ref{fig:pipeline}, the framework of \ours~primarily consists of an image encoder $\mathcal{I}_{v}$, a depth encoder $\mathcal{I}_{d}$, a vision-to-language projector $\mathcal{P}_{v-l}$, a VFE module, and an LLM. Given an underwater image $\bm{x}$, the image encoder extracts vision features $\bm{v} = \mathcal{I}_{v}\left ( \bm{x} \right )$. While employing LLaVA-1.5~\cite{liu2024improved} as the baseline, we employ a CLIP ViT-L/14~\cite{radford2021learning} with a base resolution of $336$ as the image encoder. The vision-to-language projector is a multi-layer perceptron aligning vision and language representations. It empowers the LLM Vicuna-v1.5~\cite{chiang2023vicuna} to reason about visual and textual information.

Motivated by the underwater imaging models~\cite{zhou2023underwater,akkaynak2018revised,nathan2024osmosis}, a physical prior illustrates that the imaging distance closely influences the degree of image degradation.
In particular, subjects farther away from the camera suffer more pronounced color degradation, deviating further from their original appearances.
Therefore, we adopt a frozen Depth Anything V2~\cite{yang2025depth} encoder to extract depth features $\bm{d} = \mathcal{I}_{d}\left ( \bm{x} \right )$ from given scenes. In addition, inspired by the dark pixel prior introduced in Sec.~\ref{sec:Preliminaries}, it is reasonable to measure the backscattering influence by analyzing the responses of dark pixels. 
In practice, as pixels are processed in a patch-wise manner, we identify the $k$-th image patch that exhibits the lowest average RGB value and treat the pixels within this patch as dark pixels.
Subsequently, we feed the vision feature, index $k$, and depth feature into the VFE module $\mathcal{M}$ to obtain an enhanced vision feature $\bm{v}_{e}=\mathcal{M}\left ( \bm{v},k,\bm{d} \right )$.
We would like to emphasize that both the original and enhanced vision features are essential for understanding underwater scenes.
Specifically, on the one hand, the degradations in the original vision feature reflect authentic underwater environments, facilitating the study of real ecosystems. On the other hand, the restored information in the enhanced vision feature reduces the adverse effects of imaging conditions, enabling reliable underwater perceptions. 
Therefore, we feed them forward in parallel. And due to the homologous representations of the two features, we utilize a shared projector to align them with the linguistic modality. This process can be formulated as follows:
\begin{equation}
\hat{\bm{v}} = \mathcal{P}_{v-l}\left ( \bm{v} \right ) ,\quad \hat{\bm{v}_{e}}= \mathcal{P}_{v-l}\left ( \bm{v}_{e} \right ),
\end{equation}
where $\hat{\bm{v}}$ and $\hat{\bm{v}_{e}}$ are aligned vision features derived from $\bm{v}$ and $\bm{v}_{e}$, respectively. Afterward, we employ the LLM to integrate user instructions and vision information, finally achieving multi-granular
underwater scene understandings from multi-task aspects.

\subsection{Vision Feature Enhancement} \label{sec:enhancement}
According to Eq.~\ref{eq:imaging}, the underwater vision feature enhancement comprises two steps: 1) removing backscattering and 2) restoring light absorptions.
We reveal the entire process in Fig.~\ref{fig:vfe} and clarify the reasons for taking them.

\textbf{Remove Backscattering.} Assuming $\bm{v}=\left \{ \bm{f}_{v,i} \right \} _{i=1}^{n} \in \mathbb{R}^{n\times d} $, where $n$ and $d$ denote the length and dimension of the vision feature, we regard the $k$-th slice $\bm{f}_{v,k}\in \mathbb{R}^{1\times d}$ as a dark token, representing the response of the dark pixels. 
Simply subtracting it in feature space would filter out responses from the backscattering. 
However, the dark token is encoded through multiple attention layers, which infuses global semantics beyond the backscattering. 
Therefore, we further isolate the global semantic responses from this token to estimate a pure backscattering intensity. 
Specifically, we employ a cross-attention layer with the vision feature as the key and value to aggregate global information into a learnable query. This query is also embedded with a global average feature, which is obtained by applying average pooling over the vision feature, guiding the learnable query to be more familiar with the global semantics. Assuming the output of this cross-attention layer as $\bm{q}\in \mathbb{R}^{1\times d}$, the backscattering responses $\bm{s}\in \mathbb{R}^{1\times d}$ can be refined into $\bm{f}_{v,k}-\bm{q}$. As the backscattering is added to the whole underwater scene, disturbing every image pixel, we remove backscattering by pixel-wise subtracting it from the vision feature. This process can be simply formulated as $\bm{v}-\bm{s}$, where $\bm{s}$ is broadcast to $\mathbb{R}^{n\times d}$, keeping the same shape as $\bm{v}$.

\input{figures/vfe}
\textbf{Restore Light Absorption.}
According to the underwater imaging model, underwater light is absorbed over imaging distances. We predict an absorption weight $\bm{W}\in \mathbb{R}^{n\times d}$ using a light-weight multilayer perceptron $\mathtt{MLP}$ with the depth feature as an input, i.e., $\bm{W}=\mathtt{MLP} \left ( \bm{d} \right ) $. Finally, we obtain the enhanced vision feature $\bm{v}_{e}$ by calculating:
\begin{equation}
\bm{v}_{e}=\left ( \bm{v}-\bm{s} \right )\oslash \exp{\left(-\bm{W}\right)},
\label{eq:feat_Jc}
\end{equation}
where $\oslash$ and $\exp{\left(\cdot\right)}$ denote element-wise division and 
exponentiation, respectively.
Eq.~\ref{eq:feat_Jc} conforms to Eq.~\ref{eq:Jc}, explicitly injecting human priors into network structure and regularizing restorations of visual responses in feature space.

%% file: figures/pipeline.tex
\begin{figure*}[t] 
\centering
\includegraphics[width=1.0\textwidth]{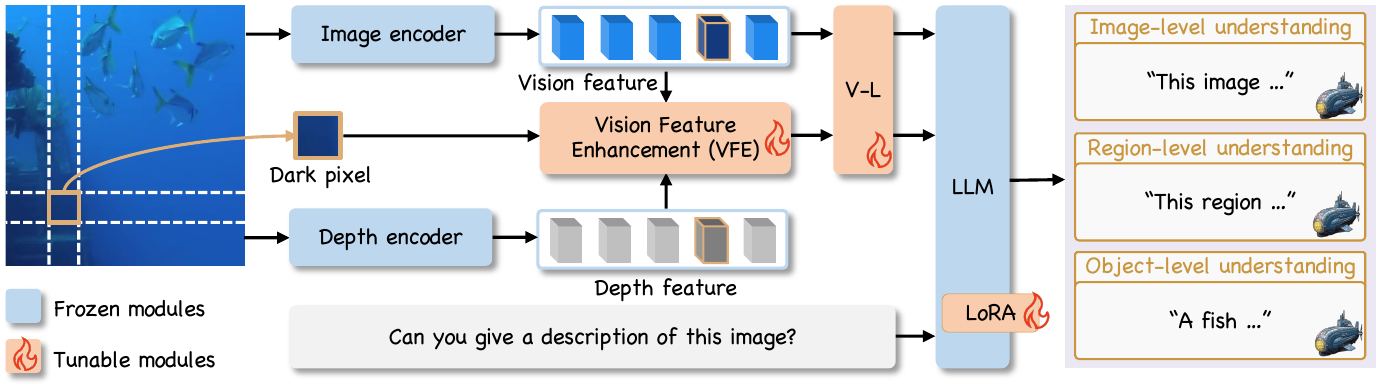}
\caption{
The framework of \ours. Inspired by underwater physical priors~\cite{zhou2023underwater,akkaynak2018revised,nathan2024osmosis}, we sample dark pixels to quantify the responses of underwater degradation. The vision feature enhancement (VFE) module improves underwater LMMs with depth information as auxiliary information. Outputs of the image encoder and the VFE module are fed into an LLM to facilitate multimodal processing.
}
\label{fig:pipeline}
\end{figure*}

%% file: figures/vfe.tex
\begin{wrapfigure}{r}{0.5\textwidth}  
\centering
\vspace{-12pt}
\includegraphics[width=0.5\textwidth]{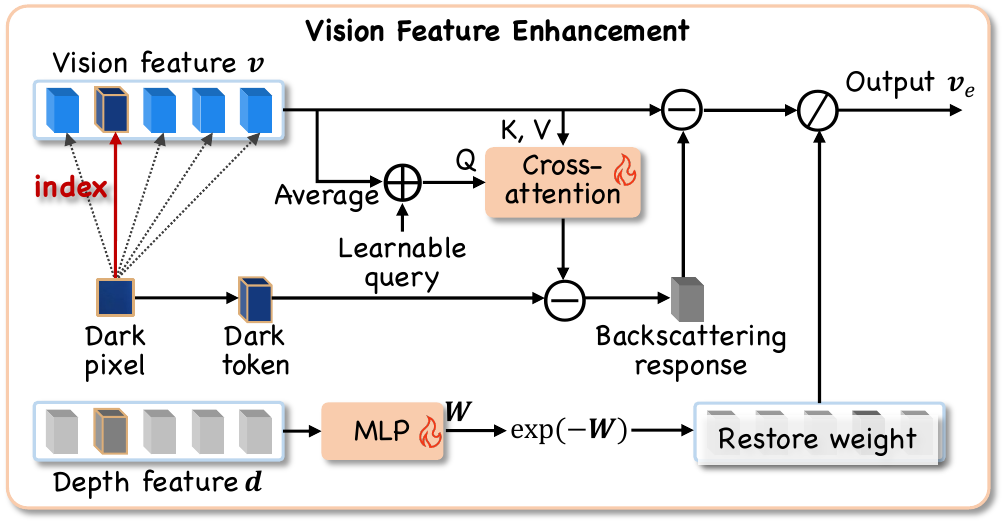}
\caption{The structure of the vision feature enhancement (VFE) module. The inputs consist of the vision feature, the index of the dark pixel, and the depth feature. It outputs enhanced vision features capturing restored underwater information.
}
\vspace{-10pt}
\label{fig:vfe}
\end{wrapfigure}

%% file: sections/experiments.tex
\section{Experiments}\label{sec:exp}

We explain the implementation details and conduct experiments to evaluate the performance of our model. As this is a pioneering work towards underwater understanding and perception employing LMMs, we select distinguished general-purpose baselines as counterparts to present comprehensive comparisons.

\textbf{Implementation Details.} We enhance vision features to address the native underwater image degradation, as described in Sec.~\ref{sec:enhancement}. To demonstrate how this design benefits current LMMs, we adapt it to LLaVA-1.5~\cite{liu2024improved} and Qwen2.5-VL~\cite{bai2025qwen2}, two prevalent LMM frameworks.
For both of them, we adopt a parameter-efficient fine-tuning (PEFT) strategy~\cite{wei2024gita,liang2024pointgst,hu2022lora}, and the trainable components are the vision-to-language projector, LoRA~\cite{hu2022lora}, and the vision feature enhancement module.
In our instruction tuning, we preserve the default hyperparameters of LLaVA-1.5 to pursue optimal performance and ensure a fair comparison with its original implementations. As for Qwen2.5-VL, since the official repository only supports full fine-tuning, we reproduce LoRA fine-tuning, setting the learning rate as $2\times 10^{-5}$ with the batch size as $16$.
The LoRA ranks in LLaVA-1.5 and Qwen2.5-VL are set as $128$.
Unless otherwise specified, we use the 7B variants, except for InternVL-2.5, for which we employ the InternVL-2.5-8B model.
Our experiments are conducted on four NVIDIA A800-80GB GPUs, training each model for one epoch, taking around $3$ days.

\textbf{Datasets.} 
\oursdata~is the first underwater instruction-following dataset providing eight-task annotations. Experiments are primarily conducted on the \oursdata. MarineInst20M~\cite{zheng2024marineinst} is a recently impressive underwater vision-language dataset containing high-quality image-caption pairs. Among this dataset, we conduct zero-shot evaluations on its human-annotated part involving the Flickr, Shutterstock, and Gettyimages subsets.
IOCfish5k~\cite{sun2023indiscernible} is the unique underwater object-counting dataset among the collected datasets. We evaluate the counting performance on its test set.

\input{tables/comparison_ft}

\subsection{Comparison to SOTA Methods}
\label{sec:comparison}

\textbf{Compared to Renowned LMMs.} 
We conduct zero-shot experiments using GPT-4o~\cite{openai2024chatml}, Qwen2.5-VL-72B~\cite{bai2025qwen2}, and Gemini 2.0 Flash~\cite{GoogleGemini2024} on the \oursbench. As shown in Tab.~\ref{tab:comparison_ft}, despite their advanced image understanding capabilities, these methods struggle to achieve satisfactory performance due to the inherent complexity of underwater environments.
Fine-tuned open-source baselines achieve significant improvements over the commercial LMMs on all tasks, demonstrating the effectiveness of domain-specific adaptation in underwater scene understanding.
The proposed \ours~explicitly addresses underwater image degradation, empowering reliable underwater perceptions under various adverse conditions.
Specifically, while employing LLaVA-1.5~\cite{liu2024improved} as the baseline, our method improves its performance on seven of eight tasks, presenting strong practical potential.
Furthermore, while employing Qwen2.5-VL~\cite{bai2025qwen2} as the base model, the \ours~ achieves the best performance on four tasks, including fine-grained classification, image caption, grounding, and detection, demonstrating remarkable underwater scene understanding capabilities.

\input{tables/comparison_ct}
\textbf{Group Perception.}
The object counting task provides insight into group behaviors. We evaluate \ours~on the underwater object counting task to assess its capability for group understanding. As shown in Tab.~\ref{tab:comparison_ct}, we follow the official data splits of IOCfish5k~\cite{sun2023indiscernible} to construct training and test subsets. Our methods outperform other LMMs by at least $0.7$ MAE and $0.3$ RMSE, delivering $8.0$ MAE and $15.9$ RMSE improvements on LLaVA-1.5, which exhibits impressive group perception performance. Nevertheless, there is a slight performance drop in the accuracy metric, which we attribute to the challenges of multi-task optimization. Specifically, the single-choice question is a text classification task substantially different from the count regression task in both objectives and output formats, introducing further challenges.

\input{tables/comparison_zs}
\textbf{Generalization.}
After fine-tuning the well-established baselines on the \oursdata, we directly evaluate their grounding performance on the MarineInst20M~\cite{zheng2024marineinst}. 
The grounding task necessitates instance-level understanding and text comprehension capabilities, which reflect the ability to achieve fine-grained understanding of underwater scenes.
As shown in Tab.~\ref{tab:comparison_zs}, our method improves the LLaVA-1.5~\cite{liu2024improved} and Qwen2.5-VL~\cite{bai2025qwen2} by $0.6$ and $0.4$ PR@0.5, respectively, demonstrating its generalization ability across domains and models.

\input{figures/visualization}

\input{tables/ablation_component}
\input{tables/ablation_condition}
\input{tables/comparison_zs_condition}

\subsection{Analysis and Ablation}
Fig.~\ref{fig:visualization} presents qualitative visualizations of \ours~across eight underwater tasks. Our model responds to user instructions and outputs multi-granularity results for underwater scene understanding, demonstrating its versatility and effectiveness. In this section, we perform ablation studies and provide an analysis of the key insights. Models in Tab.~\ref{tab:ablation_component} and Tab.~\ref{tab:ablation_condition} are trained on one-third of the \oursdata~training set and evaluated on the full \oursdata~test set. Tab.~\ref{tab:comparison_zs_condition} presents the performance of models fine-tuned on the complete \oursdata~training set when evaluated under various degraded conditions.

\textbf{Design of the VFE Module.} As shown in Tab.~\ref{tab:ablation_component}, we evaluate the effectiveness of our components. The depth encoder provides rich depth information, which is expected to enhance the LMM’s understanding of underwater scenes. However, due to the discrepancy in feature distribution, simply adding a depth encoder results in performance degradation on four tasks. In contrast, we utilize depth information to enhance light absorption and achieve improvements in five tasks compared to the second line, emphasizing the feasibility of feature fusion guided by the underwater imaging model. Subsequently, we remove backscattering to complete the restoration phase, which further surpasses the third line on five tasks, highlighting the benefits of each component.

\textbf{Enhancement in Feature Space.} We train the Qwen2.5-VL baseline with all images enhanced by Reti-Diff~\cite{he2023reti}, SMDR-IS~\cite{zhang2024synergistic}, and CCL-Net~\cite{liu2024underwater}, three state-of-the-art underwater image restoration methods, to assess the benefits of image enhancement. As shown in Tab.~\ref{tab:ablation_condition}, there are consistent performance drops in the coarse-grained classification, image caption, region caption, and detection tasks. We attribute these inferior results to the information loss introduced by image enhancement during image pre-processing. In contrast, feature enhancement preserves the image's original information to the greatest extent, leading to higher reliability and thus deserves wide application.

\textbf{Evaluation under Degraded Conditions.} Underwater environments often exhibit distinct characteristics in lighting and turbidity. To evaluate the robustness of \ours~under such domain shifts, we assess its performance across varying environmental conditions. Specifically, we employ Gemini 2.5 Flash~\cite{comanici2025gemini} to categorize the \oursdata~test set based on lighting conditions (low-light, normal-light), water turbidity (turbid, clear), and color casts (green-tinted, blue-tinted). As shown in Tab.~\ref{tab:comparison_zs_condition}, \ours~demonstrates exceptional robustness, particularly under challenging conditions. Compared to the baseline LLaVA-1.5, \ours~achieves substantial improvements of 7.5, 8.3, and 8.1 PR@0.5 in low-light, green-tinted, and turbid scenarios, respectively. Even under less challenging conditions, \ours~maintains consistent performance gains. Compared to the baseline Qwen2.5-VL, equipped with strong grounding performance, \ours~still achieves notable improvements for at least 1.2 PR@0.5 facing degradations. These results demonstrate the remarkable robustness and practical applicability of \ours~across diverse underwater environments.

%% file: tables/comparison_ft.tex
\setlength\tabcolsep{1.pt}
\begin{table*}[t]
\scriptsize
\definecolor{gainpink}{RGB}{220,0,100}
\definecolor{gaingreen}{RGB}{0,150,0}
\newcommand{\gainpink}[1]{\textcolor{gainpink}{\scalebox{0.75}{#1}}}
\newcommand{\gaingreen}[1]{\textcolor{gaingreen}{\scalebox{0.75}{#1}}}
\caption{The comparison of our \ours~and renowned LMMs on the \oursbench. The best results are highlighted in \textbf{bold}, and the second-best results are \underline{underscored}.}
\begin{tabular*}{\linewidth}{@{\extracolsep{\fill}} lcccccccccc }
\toprule
\multirow{4}{*}{Methods} & \multirow{4}{*}{Reference} & \multicolumn{2}{c}{Classification} & \multicolumn{2}{c}{Caption} & \multicolumn{2}{c}{\multirow{2.5}{*}{Grounding}} & \multicolumn{2}{c}{\multirow{2.5}{*}{Detection}} & \multicolumn{1}{c}{\multirow{2.5}{*}{VQA}}\\
\cmidrule{3-4}\cmidrule{5-6}
 & & \multicolumn{1}{c}{Coarse} & \multicolumn{1}{c}{Fine} & \multicolumn{1}{c}{Image} & \multicolumn{1}{c}{Region} &  \\ 
\cmidrule{3-11}
 & & acc$\,\uparrow$ & acc$\,\uparrow$ & METEOR$\,\uparrow$ & METEOR$\,\uparrow$ & mIoU$\,\uparrow$ & PR@0.5$\,\uparrow$ & mAP$\,\uparrow$ & mAP@0.5$\,\uparrow$ & METEOR$\,\uparrow$\\ 
\midrule
\rowcolor{gray!20} \multicolumn{11}{c}{\textit{Zero-shot experiments}} \\
\midrule
GPT-4o~\cite{openai2024chatml} & - & 55.2 &54.4 &0.179 & 0.148 & 14.2 & 4.3 & 0.3 & 1.4 & 0.242 \\ 
Qwen2.5-VL-72B~\cite{bai2025qwen2} & - & 55.2 &54.2 &0.171 & 0.126 & 42.3 & 46.4 & 8.8& 14.7 & 0.222 \\ 
Gemini 2.0 Flash~\cite{GoogleGemini2024} & - & 55.5 &54.3 &0.185 & 0.141 & 21.2 & 20.6 & 2.4 & 7.8 & 0.223 \\ 
\midrule
\rowcolor{gray!20} \multicolumn{11}{c}{\textit{Instruction-tuning experiments}} \\
\midrule
MiniGPTv2~\cite{chen2023minigpt} & arXiv 23 &80.0 &90.0 &0.204 & 0.178 & 47.0 & 51.0 & 6.9 & 12.9& 0.372 \\ 
mPLUG-Owl3~\cite{ye2024mplug3} & arXiv 24 &\textbf{91.9} &\underline{92.0} &0.219 & \textbf{0.207} & 41.1& 45.7 & 10.3 & 23.3& \textbf{0.383} \\ 
InternVL-2.5~\cite{chen2024expanding} & arXiv 24 & \underline{91.3} & 90.4 & 0.208 & 0.195 & 50.4 & 54.6 & 18.3 & 30.5 & \underline{0.382}\\ 
LLaVA-1.5~\cite{liu2024improved} & CVPR 24 &90.0 &89.8 &0.208 & 0.189 & 43.5 & 48.2 & 9.8 & 19.0& 0.359 \\ 
Qwen2.5-VL~\cite{bai2025qwen2} & arXiv 25 & 85.3 & 88.2 & \underline{0.222} & 0.196	 & \underline{52.5} & \underline{57.6}	 & \underline{24.5} & \underline{41.7}	& 0.380\\ 
\midrule
\ours\modelvariant{(LLaVA-1.5)} & - & 91.0\gainpink{(+1.0)} &89.9\gainpink{(+0.1)} & 0.208\gainpink{(+0.000)} & 0.191\gainpink{(+0.002)} & 46.2\gainpink{(+2.7)}	 & 52.2\gainpink{(+4.0)} & 11.1\gainpink{(+1.3)} & 20.9\gainpink{(+1.9)} & 0.365\gainpink{(+0.006)}\\
\ours\modelvariant{(Qwen2.5-VL)} & - & 90.3\gainpink{(+5.0)} & \textbf{93.8}\gainpink{(+5.6)} & \textbf{0.223}\gainpink{(+0.001)} & \underline{0.199}\gainpink{(+0.003)} & \textbf{53.8}\gainpink{(+1.3)} &	\textbf{58.8}\gainpink{(+1.2)} & \textbf{25.8}\gainpink{(+1.3)} & \textbf{45.3}\gainpink{(+3.6)} & 0.381\gainpink{(+0.001)}\\ 
\bottomrule
\end{tabular*}
\label{tab:comparison_ft}
\end{table*}

%% file: tables/comparison_ct.tex
\setlength\tabcolsep{2.45pt}
\begin{wraptable}{r}{0.5\textwidth}
\scriptsize
\definecolor{gainpink}{RGB}{220,0,100}
\definecolor{gaingreen}{RGB}{0,150,0}
\newcommand{\gainpink}[1]{\textcolor{gainpink}{\scalebox{0.75}{#1}}}
\newcommand{\gaingreen}[1]{\textcolor{gaingreen}{\scalebox{0.75}{#1}}}
\captionsetup{type=table}
\caption{Counting accuracy on the IOCfish5k~\cite{sun2023indiscernible} test set.}
\begin{tabular}{ lcccc }
\toprule
Methods & Reference & MAE$\,\downarrow$ & RMSE$\,\downarrow$ & acc$\,\uparrow$ \\
\midrule
\rowcolor{gray!20} \multicolumn{5}{c}{\textit{Zero-shot experiments}} \\
\midrule
GPT-4o~\cite{openai2024chatml} & - & 51.2 & \underline{94.0} & 55.4 \\
Gemini 2.0 Flash~\cite{GoogleGemini2024} & - & 50.2 & 123.3 & 50.7 \\
Qwen2.5-VL-72B~\cite{bai2025qwen2}& - & 49.8 & 124.7 & 58.8 \\
\midrule
\rowcolor{gray!20} \multicolumn{5}{c}{\textit{Instruction-tuning experiments}} \\
\midrule
MiniGPTv2~\cite{chen2023minigpt} & arXiv 23 & 55.0 & 139.6 & 43.0 \\
mPLUG-Owl3~\cite{ye2024mplug3} & arXiv 24 & 36.2 & 102.9 & 70.5 \\
InternVL-2.5~\cite{chen2024expanding} & arXiv 24 & 39.1 & 95.0 & 69.2 \\ 
LLaVA-1.5~\cite{liu2024improved} & CVPR 24 & 59.2 & 151.9 & 61.7 \\
Qwen2.5-VL~\cite{bai2025qwen2} & arXiv 25 & \underline{31.6} & 96.7 & \textbf{70.8} \\
\midrule
\ours\modelvariant{(LLaVA-1.5)} & - & 51.2\gainpink{(+8.0)} & 136.0\gainpink{(+15.9)} & 62.6\gainpink{(+0.9)} \\
\ours\modelvariant{(Qwen2.5-VL)} & - & \textbf{30.9}\gainpink{(+0.7)} &  \textbf{93.7}\gainpink{(+3.0)} & \underline{70.7}\gaingreen{(-0.1)}\\
\bottomrule
\end{tabular}
\label{tab:comparison_ct} 
\end{wraptable}

%% file: tables/comparison_zs.tex
\setlength\tabcolsep{6.7pt}
\begin{wraptable}{r}{0.5\textwidth}
\scriptsize
\definecolor{gainpink}{RGB}{220,0,100}
\definecolor{gaingreen}{RGB}{0,150,0}
\newcommand{\gainpink}[1]{\textcolor{gainpink}{\scalebox{0.75}{#1}}}
\newcommand{\gaingreen}[1]{\textcolor{gaingreen}{\scalebox{0.75}{#1}}}
\captionsetup{type=table}
\caption{Zero-shot grounding on the MarineInst20M~\cite{zheng2024marineinst}.}
\begin{tabular}{ lccc }
\toprule
Methods & Reference & mIoU$\,\uparrow$ & PR@0.5$\,\uparrow$ \\
\midrule
MiniGPTv2~\cite{chen2023minigpt} & arXiv 23 & 36.6 & 38.3\\
mPLUG-Owl3~\cite{ye2024mplug3} & arXiv 24 & 32.9 & 36.7\\
InternVL-2.5~\cite{chen2024expanding} & arXiv 24 & 37.6 & 38.4\\
LLaVA-1.5~\cite{liu2024improved} & CVPR 24 & 29.6 & 33.6\\
Qwen2.5-VL~\cite{bai2025qwen2} & arXiv 25 & \underline{38.0}	 & \underline{39.2}\\
\midrule
\ours\modelvariant{(LLaVA-1.5)} & - & 30.7\gainpink{(+1.1)} & 34.2\gainpink{(+0.6)}\\
\ours\modelvariant{(Qwen2.5-VL)} & - & \textbf{38.3}\gainpink{(+0.3)} & \textbf{39.6}\gainpink{(+0.4)}\\
\bottomrule
\end{tabular}
\label{tab:comparison_zs}
\end{wraptable}

%% file: figures/visualization.tex
\begin{figure*}[t] 
\centering
\includegraphics[width=1.0\textwidth]{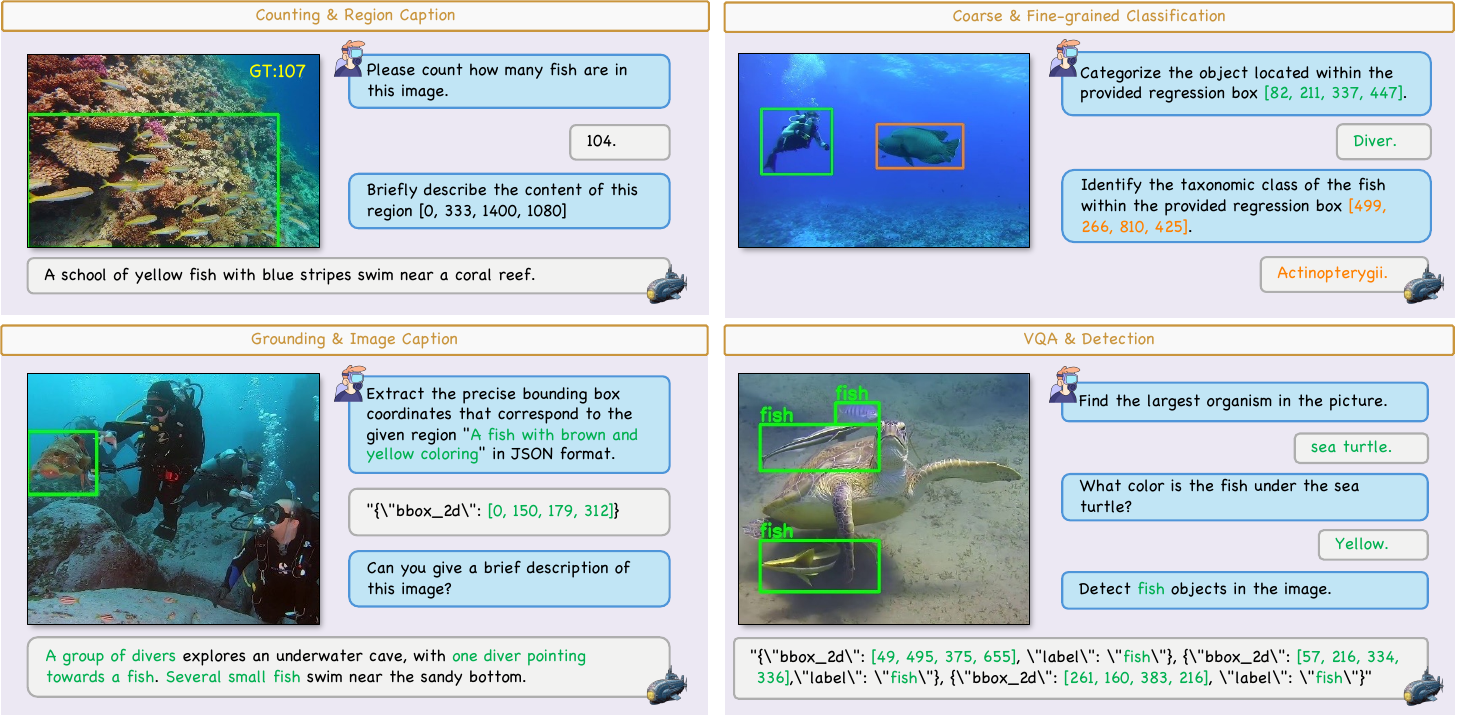}
\caption{Qualitative results on underwater scene understanding. \ours~perceives image-, region-, and object-level information while addressing eight tasks. Our underwater LMM exhibits remarkable multimodal instruction-following performance, serving as a meaningful contribution to this field.
}
\label{fig:visualization}
\end{figure*}

%% file: tables/ablation_component.tex
\setlength\tabcolsep{0.1pt}
\begin{table*}[t]
\scriptsize
\caption{Ablation on components. We employ Qwen2.5-VL~\cite{bai2025qwen2} as the baseline model and sequentially add our modules. Evaluations are conducted on eight tasks to provide a comprehensive analysis.}
\begin{tabular*}{\linewidth}{@{\extracolsep{\fill}} cccccccccccc }
\toprule
\multirow{4}{*}{Baseline} & \multirow{4}{*}{\shortstack{Depth\\encoder}} & \multicolumn{2}{c}{Vision Feature Enhancement} & \multicolumn{2}{c}{Classification} & \multicolumn{2}{c}{Caption} & \multicolumn{1}{c}{\multirow{2.5}{*}{Counting}} & \multicolumn{1}{c}{\multirow{2.5}{*}{Grounding}} & \multicolumn{1}{c}{\multirow{2.5}{*}{Detection}} & \multicolumn{1}{c}{\multirow{2.5}{*}{VQA}}\\
\cmidrule{3-4}\cmidrule{5-6}\cmidrule{7-8}
 &  &  \multirow{2.5}{*}{\shortstack{Restore light\\absorption}} & \multirow{2.5}{*}{\shortstack{Remove\\backscattering}} & \multicolumn{1}{c}{Coarse} & \multicolumn{1}{c}{Fine} & \multicolumn{1}{c}{Image} & \multicolumn{1}{c}{Region}  \\ 
\cmidrule{5-12}
& & & & acc$\,\uparrow$ & acc$\,\uparrow$ & METEOR$\,\uparrow$ & METEOR$\,\uparrow$ & MAE$\,\downarrow$ & PR@0.5$\,\uparrow$ & AP@0.5$\,\uparrow$ & METEOR$\,\uparrow$ \\ 
\midrule
\ding{52} & - & - & -  & 87.9 & 89.1 & \textbf{0.222} & \underline{0.197} & 36.8 & \underline{55.4}  & 35.9 & 0.367\\
\ding{52} & \ding{52} & - & - & \underline{89.5} & 89.1 & 0.218	& 0.194 &37.2 & 55.0 & \textbf{36.4} & 0.369 \\
\ding{52} & \ding{52} & \ding{52} & - &85.7	& \underline{91.2} & \underline{0.220} & 0.195 & \textbf{36.2} & 53.9 & 34.2 & \underline{0.372} \\
\ding{52} & \ding{52} & \ding{52} & \ding{52} & \textbf{90.0} & \textbf{91.4} & \textbf{0.222} & \textbf{0.198} & \underline{36.4} & \textbf{55.9} & \underline{36.2} & \textbf{0.373}\\
\bottomrule
\end{tabular*}
\label{tab:ablation_component}
\end{table*}

%% file: tables/ablation_condition.tex
\setlength\tabcolsep{0.1pt}
\begin{table*}[t]
\scriptsize
\caption{Ablation on components. We employ Qwen2.5-VL~\cite{bai2025qwen2} as the baseline model and sequentially add our modules. Evaluations are conducted on eight tasks to provide a comprehensive analysis.}
\begin{tabular*}{\linewidth}{@{\extracolsep{\fill}} lcccccccc }
\toprule
\multirow{4}{*}{Methods}  & \multicolumn{2}{c}{Classification} & \multicolumn{2}{c}{Caption} & \multicolumn{1}{c}{\multirow{2.5}{*}{Counting}} & \multicolumn{1}{c}{\multirow{2.5}{*}{Grounding}} & \multicolumn{1}{c}{\multirow{2.5}{*}{Detection}} & \multicolumn{1}{c}{\multirow{2.5}{*}{VQA}}\\
\cmidrule{2-3}\cmidrule{4-5}
& \multicolumn{1}{c}{Coarse} & \multicolumn{1}{c}{Fine} & \multicolumn{1}{c}{Image} & \multicolumn{1}{c}{Region}  \\ 
\cmidrule{2-9}
& acc$\,\uparrow$ & acc$\,\uparrow$ & METEOR$\,\uparrow$ & METEOR$\,\uparrow$ & MAE$\,\downarrow$ & PR@0.5$\,\uparrow$ & AP@0.5$\,\uparrow$ & METEOR$\,\uparrow$ \\ 
\midrule
Baseline &  \underline{87.9} & 89.1 & \textbf{0.222} & \underline{0.197} & 36.8 & 55.4  & \underline{35.9} & 0.367\\
+Reti-Diff~\cite{he2023reti} & 87.3 & \underline{91.1}& \underline{0.221} & 0.194 & \textbf{36.3} & \underline{55.5} & 35.0 & 0.370\\
+SMDR-IS~\cite{zhang2024synergistic} & 86.8 & 86.3 & 0.220 & 0.195 & 36.5 & 54.4 & 31.4 & \underline{0.371}\\
+CCL-Net~\cite{liu2024underwater} & 82.5 & 87.3 & 0.220 & 0.193 & 37.8 & 54.2 & 32.9 & 0.365\\
+VFE~(\textbf{ours}) & \textbf{90.0} & \textbf{91.4} & \textbf{0.222} & \textbf{0.198} & \underline{36.4} & \textbf{55.9} & \textbf{36.2} & \textbf{0.373}\\
\bottomrule
\end{tabular*}
\label{tab:ablation_condition}
\end{table*}

%% file: tables/comparison_zs_condition.tex
\setlength\tabcolsep{6.pt}
\begin{table*}[t]
\scriptsize
\definecolor{gainpink}{RGB}{220,0,100}
\definecolor{gaingreen}{RGB}{0,150,0}
\newcommand{\gainpink}[1]{\textcolor{gainpink}{\scalebox{0.75}{#1}}}
\newcommand{\gaingreen}[1]{\textcolor{gaingreen}{\scalebox{0.75}{#1}}}
\captionsetup{type=table}
\caption{Ablation on degraded conditions. We divide the \oursbench~into subsets based on the degradations of data and evaluate grounding performance on these subsets using the PR@0.5 metric.}
\begin{tabular*}{\linewidth}{@{\extracolsep{\fill}} lccccccc }
\toprule
Methods & Reference & Low-light & Normal-light & Green-tinted & Blue-tinted & Turbid & Clear \\
\midrule
MiniGPTv2~\cite{chen2023minigpt} & arXiv 23 & 44.7 & 52.8 & 47.0 & 52.7 & 46.6 & 53.5 \\
mPLUG-Owl3~\cite{ye2024mplug3} & arXiv 24 & 41.0 & 46.8 & 42.7 & 46.7 & 41.2 & 48.0\\
InternVL-2.5~\cite{chen2024expanding} & arXiv 24 & 52.7 & 55.1 & 53.4 & 55.0 & 52.7 & 55.6 \\
LLaVA-1.5~\cite{liu2024improved} & CVPR 24 & 44.1 & 49.1 & 43.5 & 50.1 & 41.9 & 51.8\\
\ours\modelvariant{(LLaVA-1.5)} & - & 51.6\gainpink{(+7.5)} & 52.0\gainpink{(+2.9)} & 51.8\gainpink{(+8.3)} & 51.9\gainpink{(+1.8)} & 50.0\gainpink{(+8.1)} & 53.1\gainpink{(+1.3)} \\
Qwen2.5-VL~\cite{bai2025qwen2} & arXiv 25 & \underline{56.9} & \underline{58.5} & \underline{56.5} & \underline{58.9} & \underline{53.4} & \underline{59.3}\\
\ours\modelvariant{(Qwen2.5-VL)} & - & \textbf{58.5}\gainpink{(+1.6)} & \textbf{58.7}\gainpink{(+0.2)} & \textbf{57.7}\gainpink{(+1.2)} & \textbf{59.2}\gainpink{(+0.3)} & \textbf{55.4}\gainpink{(+2.0)} & \textbf{60.8}\gainpink{(+1.5)}\\
\bottomrule
\end{tabular*}
\label{tab:comparison_zs_condition}
\end{table*}

%% file: sections/conclusion.tex
\section{Conclusion}\label{sec:con}
We introduce physical principles to regularize the model architecture and propose a vision feature enhancement (VFE) module. Integrating this module into renowned LLaVA-1.5 and Qwen2.5-VL, we develop \ours, the first underwater LMM addressing underwater image degradation explicitly.
Furthermore, we construct \oursdata~to bridge this gap of the absent underwater multi-task instruction-following dataset. Experiments on both \oursdata~and public underwater benchmarks demonstrate the effectiveness of the VFE module, consistently improving baselines across almost all tasks. Comparisons with current state-of-the-art methods further highlight the superiority of \ours, establishing our work as a valuable contribution to the community.

\mysection{Limitation}\label{sec:limit}
The vast diversity of underwater environments and species poses substantial challenges for exhaustively representing all relevant categories and scenarios in current datasets. Therefore, underwater scene understanding algorithms must possess open-vocabulary or few-shot learning capabilities to generalize effectively to novel and unseen cases, which is under-explored in our work.

%% file: sections/checklist.tex
\newpage
\section*{NeurIPS Paper Checklist}

The checklist is designed to encourage best practices for responsible machine learning research, addressing issues of reproducibility, transparency, research ethics, and societal impact. Do not remove the checklist: {\bf The papers not including the checklist will be desk rejected.} The checklist should follow the references and follow the (optional) supplemental material.  The checklist does NOT count towards the page
limit. 

Please read the checklist guidelines carefully for information on how to answer these questions. For each question in the checklist:
\begin{itemize}
    \item You should answer \answerYes{}, \answerNo{}, or \answerNA{}.
    \item \answerNA{} means either that the question is Not Applicable for that particular paper or the relevant information is Not Available.
    \item Please provide a short (1–2 sentence) justification right after your answer (even for NA). 
\end{itemize}

{\bf The checklist answers are an integral part of your paper submission.} They are visible to the reviewers, area chairs, senior area chairs, and ethics reviewers. You will be asked to also include it (after eventual revisions) with the final version of your paper, and its final version will be published with the paper.

The reviewers of your paper will be asked to use the checklist as one of the factors in their evaluation. While "\answerYes{}" is generally preferable to "\answerNo{}", it is perfectly acceptable to answer "\answerNo{}" provided a proper justification is given (e.g., "error bars are not reported because it would be too computationally expensive" or "we were unable to find the license for the dataset we used"). In general, answering "\answerNo{}" or "\answerNA{}" is not grounds for rejection. While the questions are phrased in a binary way, we acknowledge that the true answer is often more nuanced, so please just use your best judgment and write a justification to elaborate. All supporting evidence can appear either in the main paper or the supplemental material, provided in appendix. If you answer \answerYes{} to a question, in the justification please point to the section(s) where related material for the question can be found.

IMPORTANT, please:
\begin{itemize}
    \item {\bf Delete this instruction block, but keep the section heading ``NeurIPS Paper Checklist"},
    \item  {\bf Keep the checklist subsection headings, questions/answers and guidelines below.}
    \item {\bf Do not modify the questions and only use the provided macros for your answers}.
\end{itemize}


\begin{enumerate}

\item {\bf Claims}
    \item[] Question: Do the main claims made in the abstract and introduction accurately reflect the paper's contributions and scope?
    \item[] Answer: \answerYes{} 
    \item[] Justification: See abstract and introduction.
    \item[] Guidelines:
    \begin{itemize}
        \item The answer NA means that the abstract and introduction do not include the claims made in the paper.
        \item The abstract and/or introduction should clearly state the claims made, including the contributions made in the paper and important assumptions and limitations. A No or NA answer to this question will not be perceived well by the reviewers. 
        \item The claims made should match theoretical and experimental results, and reflect how much the results can be expected to generalize to other settings. 
        \item It is fine to include aspirational goals as motivation as long as it is clear that these goals are not attained by the paper. 
    \end{itemize}

\item {\bf Limitations}
    \item[] Question: Does the paper discuss the limitations of the work performed by the authors?
    \item[] Answer: \answerYes{} 
    \item[] Justification: See limitation part.
    \item[] Guidelines:
    \begin{itemize}
        \item The answer NA means that the paper has no limitation while the answer No means that the paper has limitations, but those are not discussed in the paper. 
        \item The authors are encouraged to create a separate "Limitations" section in their paper.
        \item The paper should point out any strong assumptions and how robust the results are to violations of these assumptions (e.g., independence assumptions, noiseless settings, model well-specification, asymptotic approximations only holding locally). The authors should reflect on how these assumptions might be violated in practice and what the implications would be.
        \item The authors should reflect on the scope of the claims made, e.g., if the approach was only tested on a few datasets or with a few runs. In general, empirical results often depend on implicit assumptions, which should be articulated.
        \item The authors should reflect on the factors that influence the performance of the approach. For example, a facial recognition algorithm may perform poorly when image resolution is low or images are taken in low lighting. Or a speech-to-text system might not be used reliably to provide closed captions for online lectures because it fails to handle technical jargon.
        \item The authors should discuss the computational efficiency of the proposed algorithms and how they scale with dataset size.
        \item If applicable, the authors should discuss possible limitations of their approach to address problems of privacy and fairness.
        \item While the authors might fear that complete honesty about limitations might be used by reviewers as grounds for rejection, a worse outcome might be that reviewers discover limitations that aren't acknowledged in the paper. The authors should use their best judgment and recognize that individual actions in favor of transparency play an important role in developing norms that preserve the integrity of the community. Reviewers will be specifically instructed to not penalize honesty concerning limitations.
    \end{itemize}

\item {\bf Theory assumptions and proofs}
    \item[] Question: For each theoretical result, does the paper provide the full set of assumptions and a complete (and correct) proof?
    \item[] Answer: \answerNA{} 
    \item[] Justification: This paper does not involve theoretical results.
    \item[] Guidelines:
    \begin{itemize}
        \item The answer NA means that the paper does not include theoretical results. 
        \item All the theorems, formulas, and proofs in the paper should be numbered and cross-referenced.
        \item All assumptions should be clearly stated or referenced in the statement of any theorems.
        \item The proofs can either appear in the main paper or the supplemental material, but if they appear in the supplemental material, the authors are encouraged to provide a short proof sketch to provide intuition. 
        \item Inversely, any informal proof provided in the core of the paper should be complemented by formal proofs provided in appendix or supplemental material.
        \item Theorems and Lemmas that the proof relies upon should be properly referenced. 
    \end{itemize}

    \item {\bf Experimental result reproducibility}
    \item[] Question: Does the paper fully disclose all the information needed to reproduce the main experimental results of the paper to the extent that it affects the main claims and/or conclusions of the paper (regardless of whether the code and data are provided or not)?
    \item[] Answer: \answerYes{} 
    \item[] Justification: See experiments part.
    \item[] Guidelines:
    \begin{itemize}
        \item The answer NA means that the paper does not include experiments.
        \item If the paper includes experiments, a No answer to this question will not be perceived well by the reviewers: Making the paper reproducible is important, regardless of whether the code and data are provided or not.
        \item If the contribution is a dataset and/or model, the authors should describe the steps taken to make their results reproducible or verifiable. 
        \item Depending on the contribution, reproducibility can be accomplished in various ways. For example, if the contribution is a novel architecture, describing the architecture fully might suffice, or if the contribution is a specific model and empirical evaluation, it may be necessary to either make it possible for others to replicate the model with the same dataset, or provide access to the model. In general. releasing code and data is often one good way to accomplish this, but reproducibility can also be provided via detailed instructions for how to replicate the results, access to a hosted model (e.g., in the case of a large language model), releasing of a model checkpoint, or other means that are appropriate to the research performed.
        \item While NeurIPS does not require releasing code, the conference does require all submissions to provide some reasonable avenue for reproducibility, which may depend on the nature of the contribution. For example
        \begin{enumerate}
            \item If the contribution is primarily a new algorithm, the paper should make it clear how to reproduce that algorithm.
            \item If the contribution is primarily a new model architecture, the paper should describe the architecture clearly and fully.
            \item If the contribution is a new model (e.g., a large language model), then there should either be a way to access this model for reproducing the results or a way to reproduce the model (e.g., with an open-source dataset or instructions for how to construct the dataset).
            \item We recognize that reproducibility may be tricky in some cases, in which case authors are welcome to describe the particular way they provide for reproducibility. In the case of closed-source models, it may be that access to the model is limited in some way (e.g., to registered users), but it should be possible for other researchers to have some path to reproducing or verifying the results.
        \end{enumerate}
    \end{itemize}

\item {\bf Open access to data and code}
    \item[] Question: Does the paper provide open access to the data and code, with sufficient instructions to faithfully reproduce the main experimental results, as described in supplemental material?
    \item[] Answer: \answerYes{} 
    \item[] Justification: The code will be made available. 
    \item[] Guidelines:
    \begin{itemize}
        \item The answer NA means that paper does not include experiments requiring code.
        \item Please see the NeurIPS code and data submission guidelines (\url{https://nips.cc/public/guides/CodeSubmissionPolicy}) for more details.
        \item While we encourage the release of code and data, we understand that this might not be possible, so “No” is an acceptable answer. Papers cannot be rejected simply for not including code, unless this is central to the contribution (e.g., for a new open-source benchmark).
        \item The instructions should contain the exact command and environment needed to run to reproduce the results. See the NeurIPS code and data submission guidelines (\url{https://nips.cc/public/guides/CodeSubmissionPolicy}) for more details.
        \item The authors should provide instructions on data access and preparation, including how to access the raw data, preprocessed data, intermediate data, and generated data, etc.
        \item The authors should provide scripts to reproduce all experimental results for the new proposed method and baselines. If only a subset of experiments are reproducible, they should state which ones are omitted from the script and why.
        \item At submission time, to preserve anonymity, the authors should release anonymized versions (if applicable).
        \item Providing as much information as possible in supplemental material (appended to the paper) is recommended, but including URLs to data and code is permitted.
    \end{itemize}

\item {\bf Experimental setting/details}
    \item[] Question: Does the paper specify all the training and test details (e.g., data splits, hyperparameters, how they were chosen, type of optimizer, etc.) necessary to understand the results?
    \item[] Answer: \answerYes{} 
    \item[] Justification: See experiments part. 
    \item[] Guidelines:
    \begin{itemize}
        \item The answer NA means that the paper does not include experiments.
        \item The experimental setting should be presented in the core of the paper to a level of detail that is necessary to appreciate the results and make sense of them.
        \item The full details can be provided either with the code, in appendix, or as supplemental material.
    \end{itemize}

\item {\bf Experiment statistical significance}
    \item[] Question: Does the paper report error bars suitably and correctly defined or other appropriate information about the statistical significance of the experiments?
    \item[] Answer: \answerYes{} 
    \item[] Justification: See experiments part. 
    \item[] Guidelines:
    \begin{itemize}
        \item The answer NA means that the paper does not include experiments.
        \item The authors should answer "Yes" if the results are accompanied by error bars, confidence intervals, or statistical significance tests, at least for the experiments that support the main claims of the paper.
        \item The factors of variability that the error bars are capturing should be clearly stated (for example, train/test split, initialization, random drawing of some parameter, or overall run with given experimental conditions).
        \item The method for calculating the error bars should be explained (closed form formula, call to a library function, bootstrap, etc.)
        \item The assumptions made should be given (e.g., Normally distributed errors).
        \item It should be clear whether the error bar is the standard deviation or the standard error of the mean.
        \item It is OK to report 1-sigma error bars, but one should state it. The authors should preferably report a 2-sigma error bar than state that they have a 96\% CI, if the hypothesis of Normality of errors is not verified.
        \item For asymmetric distributions, the authors should be careful not to show in tables or figures symmetric error bars that would yield results that are out of range (e.g. negative error rates).
        \item If error bars are reported in tables or plots, The authors should explain in the text how they were calculated and reference the corresponding figures or tables in the text.
    \end{itemize}

\item {\bf Experiments compute resources}
    \item[] Question: For each experiment, does the paper provide sufficient information on the computer resources (type of compute workers, memory, time of execution) needed to reproduce the experiments?
    \item[] Answer: \answerYes{} 
    \item[] Justification: See experiments part. 
    \item[] Guidelines:
    \begin{itemize}
        \item The answer NA means that the paper does not include experiments.
        \item The paper should indicate the type of compute workers CPU or GPU, internal cluster, or cloud provider, including relevant memory and storage.
        \item The paper should provide the amount of compute required for each of the individual experimental runs as well as estimate the total compute. 
        \item The paper should disclose whether the full research project required more compute than the experiments reported in the paper (e.g., preliminary or failed experiments that didn't make it into the paper). 
    \end{itemize}
    
\item {\bf Code of ethics}
    \item[] Question: Does the research conducted in the paper conform, in every respect, with the NeurIPS Code of Ethics \url{https://neurips.cc/public/EthicsGuidelines}?
    \item[] Answer: \answerYes{} 
    \item[] Justification: We follow the NeurIPS Code of Ethic.
    \item[] Guidelines:
    \begin{itemize}
        \item The answer NA means that the authors have not reviewed the NeurIPS Code of Ethics.
        \item If the authors answer No, they should explain the special circumstances that require a deviation from the Code of Ethics.
        \item The authors should make sure to preserve anonymity (e.g., if there is a special consideration due to laws or regulations in their jurisdiction).
    \end{itemize}

\item {\bf Broader impacts}
    \item[] Question: Does the paper discuss both potential positive societal impacts and negative societal impacts of the work performed?
    \item[] Answer: \answerNA{} 
    \item[] Justification: no societal impacts. 
    \item[] Guidelines:
    \begin{itemize}
        \item The answer NA means that there is no societal impact of the work performed.
        \item If the authors answer NA or No, they should explain why their work has no societal impact or why the paper does not address societal impact.
        \item Examples of negative societal impacts include potential malicious or unintended uses (e.g., disinformation, generating fake profiles, surveillance), fairness considerations (e.g., deployment of technologies that could make decisions that unfairly impact specific groups), privacy considerations, and security considerations.
        \item The conference expects that many papers will be foundational research and not tied to particular applications, let alone deployments. However, if there is a direct path to any negative applications, the authors should point it out. For example, it is legitimate to point out that an improvement in the quality of generative models could be used to generate deepfakes for disinformation. On the other hand, it is not needed to point out that a generic algorithm for optimizing neural networks could enable people to train models that generate Deepfakes faster.
        \item The authors should consider possible harms that could arise when the technology is being used as intended and functioning correctly, harms that could arise when the technology is being used as intended but gives incorrect results, and harms following from (intentional or unintentional) misuse of the technology.
        \item If there are negative societal impacts, the authors could also discuss possible mitigation strategies (e.g., gated release of models, providing defenses in addition to attacks, mechanisms for monitoring misuse, mechanisms to monitor how a system learns from feedback over time, improving the efficiency and accessibility of ML).
    \end{itemize}
    
\item {\bf Safeguards}
    \item[] Question: Does the paper describe safeguards that have been put in place for responsible release of data or models that have a high risk for misuse (e.g., pretrained language models, image generators, or scraped datasets)?
    \item[] Answer: \answerNA{} 
    \item[] Justification: No such risks.
    \item[] Guidelines:
    \begin{itemize}
        \item The answer NA means that the paper poses no such risks.
        \item Released models that have a high risk for misuse or dual-use should be released with necessary safeguards to allow for controlled use of the model, for example by requiring that users adhere to usage guidelines or restrictions to access the model or implementing safety filters. 
        \item Datasets that have been scraped from the Internet could pose safety risks. The authors should describe how they avoided releasing unsafe images.
        \item We recognize that providing effective safeguards is challenging, and many papers do not require this, but we encourage authors to take this into account and make a best faith effort.
    \end{itemize}

\item {\bf Licenses for existing assets}
    \item[] Question: Are the creators or original owners of assets (e.g., code, data, models), used in the paper, properly credited and are the license and terms of use explicitly mentioned and properly respected?
    \item[] Answer: \answerYes{} 
    \item[] Justification: We will release the code. 
    \item[] Guidelines:
    \begin{itemize}
        \item The answer NA means that the paper does not use existing assets.
        \item The authors should cite the original paper that produced the code package or dataset.
        \item The authors should state which version of the asset is used and, if possible, include a URL.
        \item The name of the license (e.g., CC-BY 4.0) should be included for each asset.
        \item For scraped data from a particular source (e.g., website), the copyright and terms of service of that source should be provided.
        \item If assets are released, the license, copyright information, and terms of use in the package should be provided. For popular datasets, \url{paperswithcode.com/datasets} has curated licenses for some datasets. Their licensing guide can help determine the license of a dataset.
        \item For existing datasets that are re-packaged, both the original license and the license of the derived asset (if it has changed) should be provided.
        \item If this information is not available online, the authors are encouraged to reach out to the asset's creators.
    \end{itemize}

\item {\bf New assets}
    \item[] Question: Are new assets introduced in the paper well documented and is the documentation provided alongside the assets?
    \item[] Answer: \answerNA{} 
    \item[] Justification: We use the public assets. 
    \item[] Guidelines:
    \begin{itemize}
        \item The answer NA means that the paper does not release new assets.
        \item Researchers should communicate the details of the dataset/code/model as part of their submissions via structured templates. This includes details about training, license, limitations, etc. 
        \item The paper should discuss whether and how consent was obtained from people whose asset is used.
        \item At submission time, remember to anonymize your assets (if applicable). You can either create an anonymized URL or include an anonymized zip file.
    \end{itemize}

\item {\bf Crowdsourcing and research with human subjects}
    \item[] Question: For crowdsourcing experiments and research with human subjects, does the paper include the full text of instructions given to participants and screenshots, if applicable, as well as details about compensation (if any)? 
    \item[] Answer: \answerNA{} 
    \item[] Justification: \answerNA{}
    \item[] Guidelines:
    \begin{itemize}
        \item The answer NA means that the paper does not involve crowdsourcing nor research with human subjects.
        \item Including this information in the supplemental material is fine, but if the main contribution of the paper involves human subjects, then as much detail as possible should be included in the main paper. 
        \item According to the NeurIPS Code of Ethics, workers involved in data collection, curation, or other labor should be paid at least the minimum wage in the country of the data collector. 
    \end{itemize}

\item {\bf Institutional review board (IRB) approvals or equivalent for research with human subjects}
    \item[] Question: Does the paper describe potential risks incurred by study participants, whether such risks were disclosed to the subjects, and whether Institutional Review Board (IRB) approvals (or an equivalent approval/review based on the requirements of your country or institution) were obtained?
    \item[] Answer: \answerNA{} 
    \item[] Justification: \answerNA{}
    \item[] Guidelines:
    \begin{itemize}
        \item The answer NA means that the paper does not involve crowdsourcing nor research with human subjects.
        \item Depending on the country in which research is conducted, IRB approval (or equivalent) may be required for any human subjects research. If you obtained IRB approval, you should clearly state this in the paper. 
        \item We recognize that the procedures for this may vary significantly between institutions and locations, and we expect authors to adhere to the NeurIPS Code of Ethics and the guidelines for their institution. 
        \item For initial submissions, do not include any information that would break anonymity (if applicable), such as the institution conducting the review.
    \end{itemize}

\item {\bf Declaration of LLM usage}
    \item[] Question: Does the paper describe the usage of LLMs if it is an important, original, or non-standard component of the core methods in this research? Note that if the LLM is used only for writing, editing, or formatting purposes and does not impact the core methodology, scientific rigorousness, or originality of the research, declaration is not required.
    \item[] Answer: \answerNA{} 
    \item[] Justification: \answerNA{}
    \item[] Guidelines:
    \begin{itemize}
        \item The answer NA means that the core method development in this research does not involve LLMs as any important, original, or non-standard components.
        \item Please refer to our LLM policy (\url{https://neurips.cc/Conferences/2025/LLM}) for what should or should not be described.
    \end{itemize}

\end{enumerate}

%% file: sections/appendix.tex
\newpage
\section*{Appendix}\appendix
\newcommand{\applabel}{Appendix\xspace}
\renewcommand{\thesection}{\Alph{section}}
\renewcommand{\thetable}{\Roman{table}}
\renewcommand{\thefigure}{\Roman{figure}}
\setcounter{section}{0}
\setcounter{table}{0}
\setcounter{figure}{0}

\label{sec:rationale}
In this appendix, we provide additional content to complement the main manuscript:
\begin{itemize}[leftmargin=1em,topsep=0pt]
\item \applabel~\ref{sec:apx_dataset}: Detailed information about the dataset, including 1) Distributions of the \oursdata. 2) Specific prompts for each task.
\item \applabel~\ref{sec:apx_results}: More results, including 1) Comparisons employing more metrics on specific tasks. 2) More ablation studies. 3) More visualizations.
\item \applabel~\ref{sec:apx_discussion}: Add discussions about our core insights.
\end{itemize}

\section{Detailed Information of the \oursdata}
\label{sec:apx_dataset}
\subsection{Dataset Characteristics}
\input{figures/distribution}

We collect source data from $10$ public underwater datasets and present the number of QA pairs generated from each dataset in Tab.~\ref{tab:distribution_QA}. The RUOD~\cite{fu2023rethinking} and Aquarium~\cite{aquarium} contribute to the most and least QA pairs, with the numbers of $326,068$ and $15,076$, respectively. The \oursdata~provides rich annotations covering $8$ perception and understanding tasks. As shown in Fig.~\ref{fig:distribution_images}, we count images involved in different tasks. The counting task merely contains images from the IOCfish5k~\cite{sun2023indiscernible} dataset, presenting a $1.0\%$ proportion of the total images. Most images are used to generate rule-based multi-modal instructions, constituting 55.8\% of all images.

\subsection{Generation Prompts}
\label{sec:apx_prompts}
We display the task-specific prompts in Tab.~\ref{tab:apx_dataset_prompt}. These prompts guide LMMs in generating descriptions at a specified granularity and in a prescribed format. The generated descriptions are subsequently combined with pre-defined questions to construct QA pairs. In particular, we provide pre-defined question templates in Tab.~\ref{tab:apx_question_template}.

\section{More Results}
\label{sec:apx_results}

\subsection{Detailed Results on Each Task}
\label{sec:apx_detailedresults}
\input{tables/X_sup_Caption_Detection}

\input{tables/X_sup_Classification_Grounding}

We perform comparisons on all supported tasks to provide a comprehensive evaluation of the proposed \ours. As shown in Tab.~\ref{tab:apx_detection} and Tab.~\ref{tab:apx_grounding}, our model ranks first on five of six metrics, highlighting its superior object-level perception capability.
Tab.~\ref{tab:apx_caption} illustrates the underwater image and region caption performance of current SOTAs. Our method presents superior performance on the image caption task, surpassing other methods on both the CIDEr and METEOR metrics. However, it struggles to achieve optimal performance on the region caption task, second to mPLUG-Owl3~\cite{ye2024mplug3}, a recent LMM renowned for its high-level image understanding performance, which also performs best on the coarse-grained classification task, demonstrated in Tab.~\ref{tab:apx_classification}. 
Nonetheless, it is worth noting that the \ours~presents impressive capabilities on the fine-grained classification task, surpassing other methods on three metrics. We attribute this to the effectiveness of the VFE module in enhancing the underwater vision feature, which empowers our models to perceive more fine-grained information.

\subsection{Ablation Study}

\input{tables/X_sup_ablation_weight}

We conduct ablation studies on the weighting strategies, with results presented in Tab.~\ref{tab:X_sup_ablation_weight}. In particular, we explore three designs of weighting strategies, including ``wo/ weighting'', ``Learned from image feature'', and ``Learned from depth feature (ours)''. Among them, the ``wo/ weighting'' strategy can be considered as ``a norm weight 1''. The ``Learned from image feature'' strategy, which means a loss of depth information, presents comparable performance compared with our design in most tasks. However, depth information is essential for distance perceptions of complex underwater objects, benefiting underwater scene understanding intuitively. Our strategy with the fusion of depth features surpasses other strategies, indicating the effectiveness of our design choice.

\input{figures/X_sup_visualization}
\subsection{Visualizations}
\label{sec:apx_visualizations}

We provide visualizations employing our \ours~in more underwater scenes, as shown in Fig.~\ref{fig:apx_visualization}. Despite the significant diversity in these scenarios, including substantial variations in lighting conditions and viewpoints, the \ours~accurately localizes the targets and provides reasonable textual descriptions. These qualitative results further illustrate the effectiveness of our method.

\section{Discussion}
\label{sec:apx_discussion}

\input{figures/case}

The underwater imaging model leads a physics-driven research direction to tackle image degradation. Explicit information restoration in feature space presents a distinct advantage for our \ours, especially in high-level understanding tasks. Specifically, image augmentation changes pixel values, resulting in compromised preservation of the image’s fidelity and semantics. In contrast, the proposed feature enhancement method incorporates physical priors to regularize the feature extraction and interaction process, preserving the original image information and demonstrating greater application potential.

As shown in Fig.~\ref{fig:apx_case}, the left and right columns show original and corresponding augmented underwater images. Feeding them into the representative multimodal large language model GPT-4o~\cite{openai2024chatml} produces markedly different feedback. In particular, while employing the data augmentation process, the GPT-4o misinterprets the \textit{dim and diffuse} lighting as \textit{nature and moderately bright}, illustrating the necessity of exploring feature enhancement in the underwater scene understanding task.

\clearpage

\input{prompts/X_sup_dataset_prompt}
\input{prompts/X_sup_question_template}

%% file: figures/distribution.tex
\begin{figure}[ht]
        \small
    \begin{minipage}{0.52\textwidth}
        \centering
        \captionsetup{type=table}
        \setlength{\tabcolsep}{1.2mm}
        \caption{Distribution of QA pairs generated from different datasets. ``Images'' indicates the number of images collected from each dataset. ``Proportion'' refers to the percentage of QA pairs derived from each dataset relative to the total number of QA pairs.
        }
        \begin{tabular}{lcccc}
            \toprule
            Dataset & Images & QA pairs & Proportion \\ 
            \midrule
            USIS10k~\cite{lian2024diving} & 10,632 & 179,772 & 12.5\% \\
            UIIS~\cite{Lian_2023_ICCV} & 4,628 & 123,492 & 8.6\% \\
            RUOD~\cite{fu2023rethinking} & 14,000 & 326,068 & 22.6\% \\
            Deepfish~\cite{saleh2020realistic} &  4,505 & 97,007 & 6.7\% \\
            Brackish~\cite{pedersen2023brackishmot} & 12,444 & 245,358 & 17.0\% \\
            IOCfish5k~\cite{sun2023indiscernible} &  5,637 & 64,037 & 4.4\% \\
            UVOT-400~\cite{alawode2023improving} & 9,064 & 169,988 & 11.8\% \\
            Aquarium~\cite{aquarium} & 638 & 15,076 & 1.0\% \\
            Underwater Trash~\cite{underwater-garbage} & 5,130 & 33,429 & 2.3\% \\
            FishNet~\cite{khan2023fishnet} & 94,806 & 188,450 & 13.1\% \\
            \bottomrule
        \end{tabular}
        \label{tab:distribution_QA} 
    \end{minipage}
    \hfill
    \begin{minipage}{0.47\textwidth}
        \centering
        \includegraphics[width=1.\textwidth]{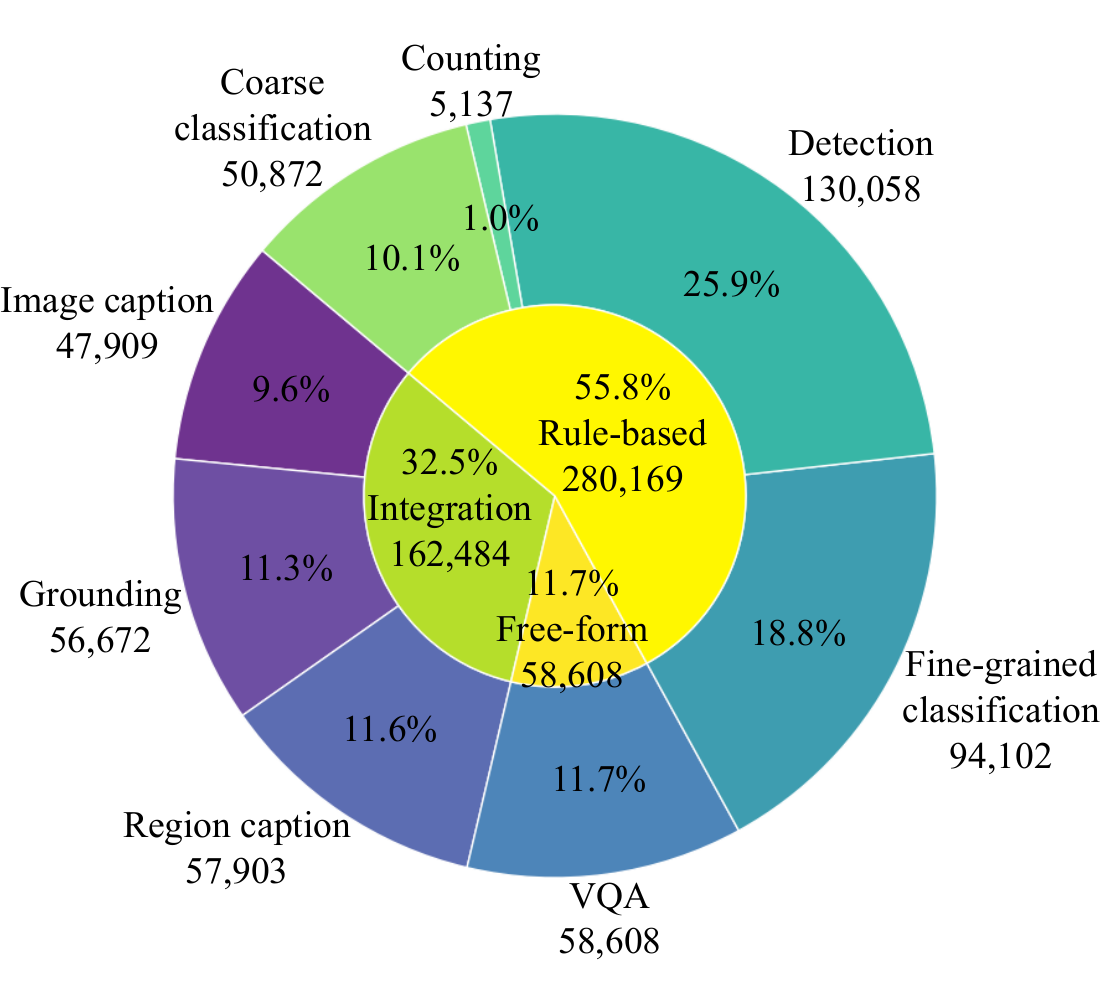}
        \caption{The distribution of images involved in different tasks.
        }
        \label{fig:distribution_images}
    \end{minipage}    
\end{figure}

%% file: tables/X_sup_Caption_Detection.tex
\begin{figure}[ht]
    \scriptsize
    \begin{minipage}{0.385\textwidth}
        \centering
        \captionsetup{type=table}
        \setlength{\tabcolsep}{0.mm}
        \caption{Comparisons on the detection task.
        }
        \begin{tabular}{lcccc}
            \toprule
            Method & mAP@0.75$\,\uparrow$ & AR@100$\,\uparrow$ & mAP$\,\uparrow$ \\ 
            \midrule
            \rowcolor{gray!20} \multicolumn{4}{c}{\textit{Zero-shot experiments}} \\
            \midrule
            GPT-4o~\cite{openai2024chatml} & 0.03 & 0.99 & 0.30   \\
            Qwen2.5-VL-72B~\cite{bai2025qwen2} & 8.58 & 14.04 & 8.79   \\
            Gemini 2.0 Flash~\cite{GoogleGemini2024} & 0.64 & 5.14 & 2.36   \\
            \midrule
            \rowcolor{gray!20} \multicolumn{4}{c}{\textit{Instruction-tuning experiments}} \\
            \midrule
            MiniGPTv2~\cite{chen2023minigpt} & 6.78 & 10.56 & 6.94   \\
            mPLUG-Owl3~\cite{ye2024mplug3} & 6.13 & 18.51 & 10.32   \\
            InternVL-2.5~\cite{chen2024expanding} & 18.11 & 29.17 & 18.28  \\
            LLaVA-1.5~\cite{liu2024improved} & 8.59 & 16.81 & 9.79   \\
            Qwen2.5-VL~\cite{bai2025qwen2} & \textbf{23.64} & \underline{33.39} & \underline{24.50}   \\
            \midrule
            \ours\modelvariant{(LLaVA-1.5)} & 10.25 & 19.37 & 11.06   \\
            \ours\modelvariant{(Qwen2.5-VL)} & \underline{23.50} & \textbf{34.88} & \textbf{25.75}   \\
            \bottomrule
        \end{tabular}
        \label{tab:apx_detection} 
    \end{minipage}
    \hfill
    \begin{minipage}{0.6\textwidth}
        \centering
        \captionsetup{type=table}
        \setlength{\tabcolsep}{0.mm}
        \caption{Comparisons on the image caption and region caption tasks.
        }
        \begin{tabular}{lcccccc}
            \toprule
            \multirow{2.5}{*}{Method} & \multicolumn{3}{c}{Image Caption} & \multicolumn{3}{c}{Region Caption} \\ 
            \cmidrule{2-4}\cmidrule{5-7}
            & BLEU-4$\,\uparrow$ & CIDEr$\,\uparrow$ & METEOR$\,\uparrow$ & BLEU-4$\,\uparrow$ & CIDEr$\,\uparrow$ & METEOR$\,\uparrow$ \\
            \midrule
            \rowcolor{gray!20} \multicolumn{7}{c}{\textit{Zero-shot experiments}} \\
            \midrule
            GPT-4o~\cite{openai2024chatml} & 0.047 & 0.322 & 0.179 & 0.018 & 0.137 & 0.148   \\
            Qwen2.5-VL-72B~\cite{bai2025qwen2} & 0.037 & 0.205 & 0.171 & 0.043 & 0.333 & 0.126   \\
            Gemini 2.0 Flash~\cite{GoogleGemini2024} & 0.050 & 0.248 & 0.185 & 0.022 & 0.146 & 0.141   \\
            \midrule
            \rowcolor{gray!20} \multicolumn{7}{c}{\textit{Instruction-tuning experiments}} \\
            \midrule
            MiniGPTv2~\cite{chen2023minigpt} & 0.107 & 0.794 & 0.204 & 0.097 & 0.792 & 0.178   \\
            mPLUG-Owl3~\cite{ye2024mplug3} & 0.130 & 0.979 & 0.219 & \textbf{0.156 }& \underline{1.245} & \textbf{0.207}   \\
            InternVL-2.5~\cite{chen2024expanding} & 0.126 & 0.924 & 0.208 & 0.147 & 1.177 & 0.195   \\
            LLaVA-1.5~\cite{liu2024improved} & 0.126 & 0.902 & 0.208 & 0.141 & 1.104 & 0.189   \\
            Qwen2.5-VL~\cite{bai2025qwen2} & \textbf{0.140} & \underline{1.015} & \underline{0.222} & \underline{0.149} & 1.230 & 0.196   \\
            \midrule
            \ours\modelvariant{(LLaVA-1.5)} & 0.132 & 0.947 & 0.208 & 0.140 & 0.109 & 0.191   \\
            \ours\modelvariant{(Qwen2.5-VL)} & \underline{0.139} & \textbf{1.023} & \textbf{0.223} & 0.148 & \textbf{1.258} & \underline{0.199}   \\
            \bottomrule
        \end{tabular}
        \label{tab:apx_caption} 
    \end{minipage}    
\end{figure}

%% file: tables/X_sup_Classification_Grounding.tex
\begin{figure}[ht]
        \scriptsize
        \begin{minipage}{0.385\textwidth}
        \centering
        \captionsetup{type=table}
        \setlength{\tabcolsep}{0.mm}
        \caption{Comparisons on the grounding task.
        }
        \begin{tabular}{lccc}
            \toprule
            Method  & AP@0.5$\,\uparrow$ & PR@0.75$\,\uparrow$ & PR@0.5$\,\uparrow$ \\ 
            \midrule
            \rowcolor{gray!20} \multicolumn{4}{c}{\textit{Zero-shot experiments}} \\
            \midrule
            GPT-4o~\cite{openai2024chatml} & 0.37 & 0.54 & 4.31 \\
            Qwen2.5-VL-72B~\cite{bai2025qwen2} & 30.90 & 33.11 & 46.36 \\
            Gemini 2.0 Flash~\cite{GoogleGemini2024} & 6.42 & 7.55 & 20.60 \\
            \midrule
            \rowcolor{gray!20} \multicolumn{4}{c}{\textit{Instruction-tuning experiments}} \\
            \midrule
            MiniGPTv2~\cite{chen2023minigpt} & 32.75 & 41.19 & 50.99 \\
            mPLUG-Owl3~\cite{ye2024mplug3} & 28.15 & 26.62 & 45.70 \\
            InternVL-2.5~\cite{chen2024expanding} & 37.50 & 44.69 & 54.64 \\
            LLaVA-1.5~\cite{liu2024improved} & 31.04 & 30.46 & 48.21 \\
            Qwen2.5-VL~\cite{bai2025qwen2} & \underline{39.34} & \underline{47.42} & \underline{57.62} \\
            \midrule
            \ours\modelvariant{(LLaVA-1.5)} & 34.19 & 36.16 & 52.19 \\
            \ours\modelvariant{(Qwen2.5-VL)} & \textbf{40.89} & \textbf{48.08} & \textbf{58.81} \\
            \bottomrule
        \end{tabular}
        \label{tab:apx_grounding} 
    \end{minipage}
    \hfill
    \begin{minipage}{0.6\textwidth}
        \centering
        \captionsetup{type=table}
        \setlength{\tabcolsep}{2.mm}
        \caption{Comparisons on the coarse-grained and fine-grained classification tasks.
        }
        \begin{tabular}{lcccccc}
            \toprule
            \multirow{2.5}{*}{Method} & \multicolumn{3}{c}{Coarse-grained} & \multicolumn{3}{c}{Fine-grained} \\ 
            \cmidrule{2-4}\cmidrule{5-7}
            & PR$\,\uparrow$ & F1$\,\uparrow$ & acc$\,\uparrow$ & PR$\,\uparrow$ & F1$\,\uparrow$ & acc$\,\uparrow$ \\
            \midrule
            \rowcolor{gray!20} \multicolumn{7}{c}{\textit{Zero-shot experiments}} \\
            \midrule
            GPT-4o~\cite{openai2024chatml} & 84.40 & 63.46 & 55.18 & 67.58 & 43.43 & 54.44   \\
            Qwen2.5-VL-72B~\cite{bai2025qwen2} & 73.37 & 60.30 & 55.18 & 54.65 & 54.43 & 54.24   \\
            Gemini 2.0 Flash~\cite{GoogleGemini2024} & 64.35 & 58.29 & 55.45 & 31.27 & 39.16 & 54.34   \\
            \midrule
            \rowcolor{gray!20} \multicolumn{7}{c}{\textit{Instruction-tuning experiments}} \\
            \midrule
            MiniGPTv2~\cite{chen2023minigpt} & 82.17 & 79.53 & 79.95 & 89.94 & 89.91 & 90.00   \\
            mPLUG-Owl3~\cite{ye2024mplug3} & \textbf{92.07} & \textbf{91.65} & \textbf{91.92} & \underline{92.13} & \underline{92.02} & \underline{92.02}   \\
            InternVL-2.5~\cite{chen2024expanding} & 91.01 & \underline{90.79} & \underline{91.25} & 90.34 & 90.33 & 90.40   \\
            LLaVA-1.5~\cite{liu2024improved} & 90.11 & 89.46 & 90.04 & 89.72 & 89.72 & 89.80   \\
            Qwen2.5-VL~\cite{bai2025qwen2} & 82.31 & 82.94 & 85.33 & 90.02 & 88.19 & 88.18   \\
            \midrule
            \ours\modelvariant{(LLaVA-1.5)} & \underline{91.06} & 90.49 & 90.98 & 89.95 & 89.86 & 89.90   \\
            \ours\modelvariant{(Qwen2.5-VL)} & 90.98 & 89.88 & 90.31 & \textbf{93.80} & \textbf{93.80} & \textbf{93.84}   \\
            \bottomrule
        \end{tabular}
        \label{tab:apx_classification} 
    \end{minipage}    
\end{figure}

%% file: tables/X_sup_ablation_weight.tex
\setlength\tabcolsep{0.1pt}
\begin{table*}[t]
\scriptsize
\caption{Ablation studies on the weighting strategies. We employ Qwen2.5-VL~\cite{bai2025qwen2} as the baseline model. Evaluations are conducted on eight tasks to provide a comprehensive analysis.}
\begin{tabular*}{\linewidth}{@{\extracolsep{\fill}} lcccccccc }
\toprule
\multirow{4}{*}{Methods}  & \multicolumn{2}{c}{Classification} & \multicolumn{2}{c}{Caption} & \multicolumn{1}{c}{\multirow{2.5}{*}{Counting}} & \multicolumn{1}{c}{\multirow{2.5}{*}{Grounding}} & \multicolumn{1}{c}{\multirow{2.5}{*}{Detection}} & \multicolumn{1}{c}{\multirow{2.5}{*}{VQA}}\\
\cmidrule{2-3}\cmidrule{4-5}
& \multicolumn{1}{c}{Coarse} & \multicolumn{1}{c}{Fine} & \multicolumn{1}{c}{Image} & \multicolumn{1}{c}{Region}  \\ 
\cmidrule{2-9}
& acc$\,\uparrow$ & acc$\,\uparrow$ & METEOR$\,\uparrow$ & METEOR$\,\uparrow$ & MAE$\,\downarrow$ & PR@0.5$\,\uparrow$ & AP@0.5$\,\uparrow$ & METEOR$\,\uparrow$ \\ 
\midrule
wo/ weighting & 87.9 & 89.1 & \textbf{0.222} & \underline{0.197} & 36.8 & 55.4 & 35.9 & 0.367 \\ 
Learned from image feature & \underline{88.6} & \underline{90.2}	& \underline{0.220}	& 0.195	& \underline{36.5} & \underline{55.7} & \textbf{36.3} & \textbf{0.375} \\
Learned from depth feature & \textbf{90.0} & \textbf{91.4} & \textbf{0.222} & \textbf{0.198} & \textbf{36.4} & \textbf{55.9} & \underline{36.2} & \underline{0.373} \\ 
\bottomrule
\end{tabular*}
\label{tab:X_sup_ablation_weight}
\end{table*}

%% file: figures/X_sup_visualization.tex
\begin{figure*}[t] 
\centering
\includegraphics[width=1.0\textwidth]{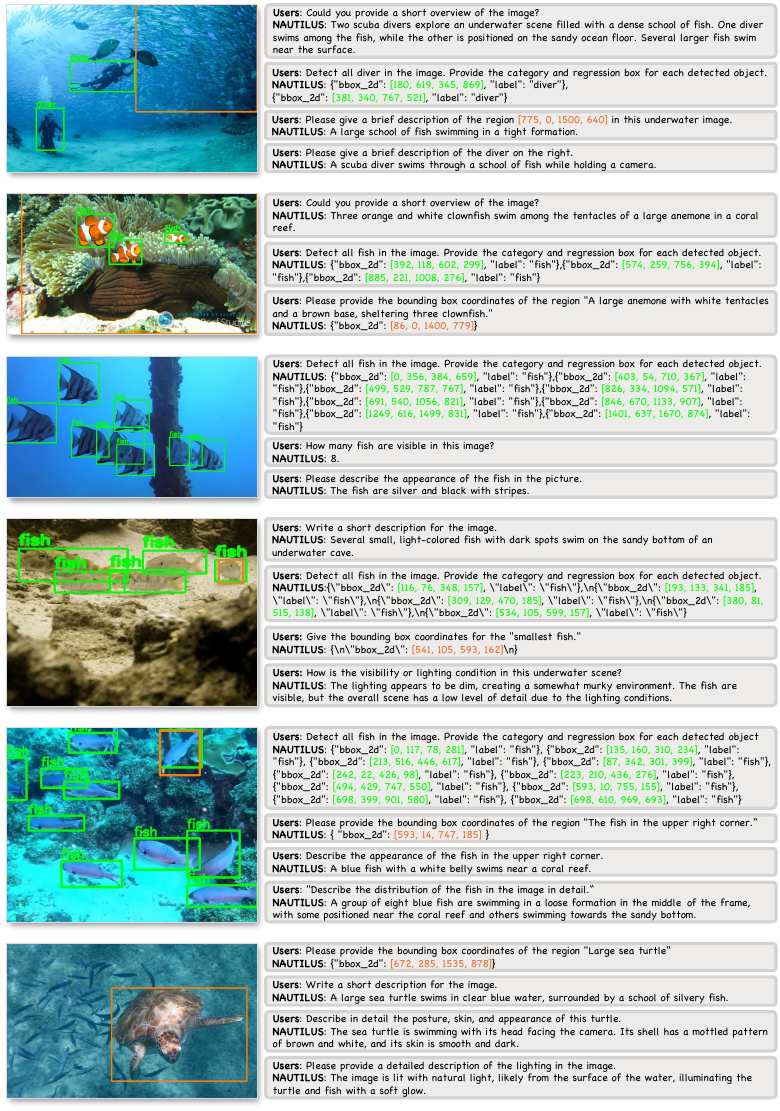}
\caption{Visualizations across various underwater scenes encompassing different illumination conditions, target categories, individual scales, environments, etc. Our \ours~consistently presents remarkable underwater scene understanding performance.
}
\label{fig:apx_visualization}
\end{figure*}

%% file: figures/case.tex
\begin{wrapfigure}{r}{0.5\textwidth} 
  \centering
  \includegraphics[width=0.49\textwidth]{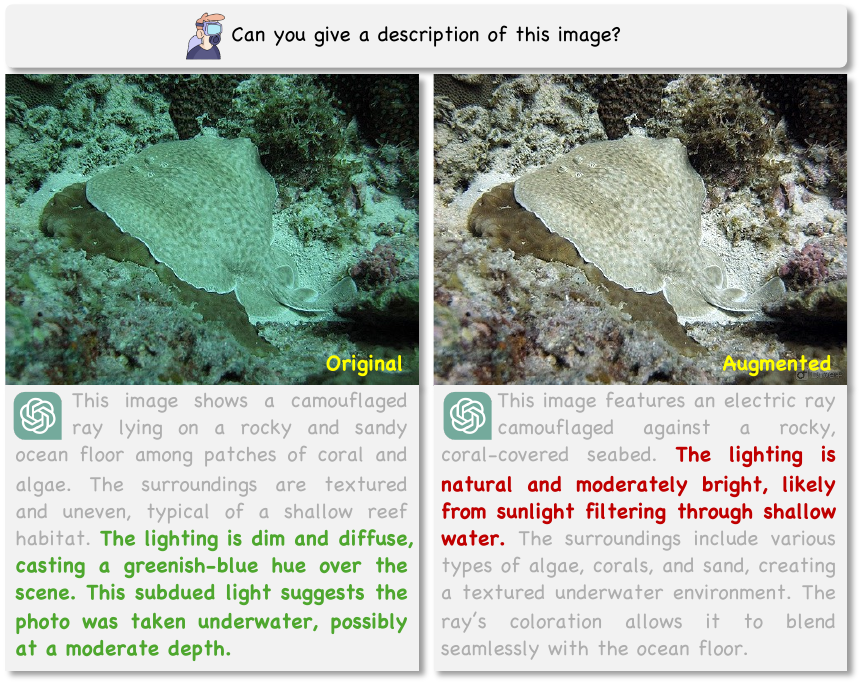}
  \caption{Feeding a pair of original and augmented underwater images into an LMM yields output descriptions with notable differences, indicating that data augmentation can change the semantic interpretation of the image.}
  \label{fig:apx_case}
\end{wrapfigure}

%% file: prompts/X_sup_dataset_prompt.tex
\begin{table*}[htbp]
\centering
\caption{Prompts designed to generate text annotations.}
\label{tab:apx_dataset_prompt}
\begin{tcolorbox}[colback=gray!10]
\centering
\footnotesize
\begin{tabular}{p{0.97\columnwidth} c}

\PromptSty{\textbf{Prompt 1: Image caption}}\\
You are an AI visual assistant analyzing an underwater image.

Task Overview:

Generate a detailed and accurate description of the image in one sentence.

Image caption guidelines:

1. Ensure that **each description is affirmative and can be inferred clearly from the image**.

2. Identify the main targets such as various fish, sea turtles, jellyfish and other marine organisms, shipwrecks and ruins, coral reefs, seagrass, divers, etc. When describing, mention the accurate number of targets and determine their category.

3. Consider the action or state of the objects, relationship between objects, background and environment.

\hrulefill \\ 

\PromptSty{\textbf{Prompt 2: Region caption}}\\

Given an image, a bounding box (bbox), and additional textual context, generate a high-quality region description. The description must adhere to the following principles:

Input:

bounding box: \{bbox\}

This bounding box represents the normalized xy\-coordinates of the top-left and bottom-right corners of the target region in the image.

Accuracy: Ensure the description precisely reflects the content within the specified bbox without adding speculative or unrelated details.

Specificity: Provide concrete details about the object's attributes (e.g., shape, color, texture) and relevant contextual information.

Objectivity: Avoid any subjective interpretations, emotions, or assumptions about the object's purpose or intent.

Conciseness: Keep the description informative yet succinct, avoiding unnecessary elaboration.

Context Awareness: Consider the surrounding elements only if they are relevant to understanding the object in the bbox.

Output Format:

description: A dark\-colored fish with a broad body and a slightly pointed head, swimming near the coral reef. \\

\hrulefill \\ 

\PromptSty{\textbf{Prompt 3: VQA}}\\
You are an AI visual assistant analyzing an underwater image. Generate a structured dialogue between yourself and a person asking questions about the image. Your task is to create precise question-answer pairs based purely on the observable visual content of the image. Each answer should be as short as possible, preferably a single word or short phrase, while maintaining accuracy.

Task Overview: \\
Generate a variety of structured question-answer pairs that reflect the image's content. Questions should cover different aspects of the image and fall into one of the following categories:
Object Recognition Questions: Identifying or detecting object types and categories (e.g., fish species, coral structures).
Attribute Questions: Describing the properties of objects (e.g., color, size, shape, material).
Counting Questions: Asking about the number of specific objects (e.g., number of fish or coral formations).
Spatial Relation Questions: Asking about the relative position or spatial layout of objects (e.g., where objects are located or their relative positions ).

Guidelines: \\
For "Spatial Relation Questions",do not answer them by "In the ocean".
Ensure that every question has a definite and clear answer based on what is visually observable in the image.
Avoid speculative or ambiguous questions. Questions should be answerable with confidence and based on visible content.
Include both simple (object identification, counting) and moderate (relative positioning, behaviors) questions.

Format: \\
Follow this exact format for each question-answer pair and no need to include other content: \\
Q: [Question] \\
A: [A single word or phrase]

\end{tabular}
\end{tcolorbox}
\end{table*}

%% file: prompts/X_sup_question_template.tex
\begin{table*}[t!]
\centering
\caption{A list of question templates employed to construct task-specific conversions.}
\label{tab:apx_question_template}
\begin{tcolorbox}[colback=gray!10]
\centering
\footnotesize
\begin{tabular}{p{0.97\columnwidth} c}

\PromptSty{\textbf{Task 1: Image caption}}\\
1. Write a description for the image. \\
2. Offer a concise description of the image. \\
3. Give a description of the scene. \\
4. Give a short description of the image. \\
5. Provide a brief description of the image. \\
6. Please give a succinct description of what is shown in this image. \\
7. Could you describe the image briefly? \\
8. Could you provide a description of the image? \\
9. Can you give a brief description of this image? \\
10. Could you provide a short overview of the image? \\
\hrulefill \\ 

\PromptSty{\textbf{Task 2: Region caption}} \\
1. Please provide a concise description of this region [bbox] in this underwater image. \\
2. Please give a brief description of the region [bbox] in this underwater image. \\
3. Could you provide a concise description of the region [bbox] in this underwater image? \\
4. Please offer a succinct description of the region [bbox]. \\
5. Please provide a short yet informative description of the region [bbox]. \\
6. Describe this region [bbox] in the underwater image. \\
7. Briefly describe the content of this region [bbox]. \\
8. In this underwater image, please provide a concise description of the region [bbox]. \\
9. For this underwater image, concisely describe the content of the region [bbox]. \\
10. What's in this region [bbox] of the underwater image? Describe it concisely. \\
\hrulefill \\ 

\PromptSty{\textbf{Task 3: Grounding}} \\
1. Please locate the bounding box coordinates of the [region]. \\
2. Find and return the bounding box coordinates of the [region]. \\
3. Give the bounding box coordinates for the [region]. \\
4. Extract the precise bounding box coordinates that correspond to the given [region]. \\
5. Detect and outline the bounding box coordinates enclosing the [region]. \\
\hrulefill \\ 

\PromptSty{\textbf{Task 4: Detection}} \\
1. Detect all [class] object in the image. \\
2. Detect all underwater object in the image, including [class1], [class2], ..., [classn]. \\
\hrulefill \\ 

\PromptSty{\textbf{Task 5: Counting}} \\
1. How many fish can you find in this image? \\  
2. Please count the fish in the image.   \\
3. Identify the number of fish present in this image.   \\
4. Count the total number of fish visible in this image.   \\
5. What is the count of fish in this image?   \\
6. How many fish can you see in this image?   \\
7. How many fish are visible in this image?   \\
8. Please count how many fish are in this image.   \\
9. Can you determine the number of fish in this image?   \\
10. What is the total number of fish shown in this image? \\
\hrulefill \\ 

\PromptSty{\textbf{Task 6: Coarse-grained classification}} \\
1. Identify the object inside the specified regression box [bbox]. \\
2. Categorize the object located within the provided regression box [bbox] in the underwater image. \\
3. Classify the items found inside the given regression box [bbox]. \\
4. Determine the category of the object inside the specified regression box [bbox]. \\
5. Assign a category to the object inside the provided regression box [bbox] in the image. \\
\hrulefill \\ 

\PromptSty{\textbf{Task 7: Fine-grained classification}} \\
1. Please identify the biological class of fish depicted in the image.  \\
2. Can you recognize the fish taxonomic class shown in this image?  \\
3. Could you determine the taxonomic class of the fish in the image?  \\
4. What is the biological class of the fish in the image?  \\
5. You are requested to identify the biological class of fish present in the image.  \\

\end{tabular}
\end{tcolorbox}
\end{table*}